\documentclass[conference]{IEEEtran}
\IEEEoverridecommandlockouts
\usepackage[utf8]{inputenc}
\usepackage{graphicx}
\usepackage{amsmath}
\usepackage[version=4]{mhchem}
\usepackage{siunitx}
\usepackage{amsmath,amsfonts}
\usepackage{subcaption}
\usepackage{mathtools}
\usepackage{booktabs}
\usepackage{multicol}
\usepackage{multirow}
\usepackage{cleveref}
\usepackage{algorithm}
\usepackage{algpseudocode}
\usepackage[backend=bibtex,style=ieee,sorting=none]{biblatex}
\usepackage{longtable,tabularx}

\begin{document}

\title{Physics-Informed Machine Learning Towards A Real-Time Spacecraft Thermal Simulator}

\makeatletter
\newcommand{\linebreakand}{%
  \end{@IEEEauthorhalign}
  \hfill\mbox{}\par
  \mbox{}\hfill\begin{@IEEEauthorhalign}
}
\makeatother

\author{
\IEEEauthorblockN{Manaswin Oddiraju \thanks{PhD Candidate, Department of Mechanical and Aerospace Engineering, AIAA Student Member}}
\IEEEauthorblockA{University at Buffalo \\Buffalo, NY, 14260}
\and
\IEEEauthorblockN{Zaki Hasnain \thanks{Data Scientist, Systems Engineering Division}}
\IEEEauthorblockA{Jet Propulsion Laboratory\\ Pasadena, CA, 91109}
\and
\IEEEauthorblockN{Saptarshi Bandyopadhyay \thanks{Robotics Technologist, Autonomous Systems Division, AIAA Member}} 
\IEEEauthorblockA{Jet Propulsion Laboratory\\ Pasadena, CA, 91109}

\linebreakand 
\IEEEauthorblockN{Eric Sunada \thanks{Senior Thermal Engineer, Mechanical Systems Engineering, Fabrication And Test Division}}
\IEEEauthorblockA{Jet Propulsion Laboratory\\ Pasadena, CA, 91109}
\and
\IEEEauthorblockN{Souma Chowdhury\thanks{Associate Professor, Department of Mechanical and Aerospace Engineering, Co-Director, Center for Embodied Autonomy and Robotics, AIAA Associate Fellow, Corresponding author. Email: soumacho@buffalo.edu}}
\IEEEauthorblockA{University at Buffalo \\ Buffalo, NY, 14260}
}

\maketitle

\begin{abstract}
Modeling thermal states for complex space missions, such as the surface exploration of airless bodies, requires high computation, whether used in ground-based analysis for spacecraft design or during onboard reasoning for autonomous operations. For example, a finite-element-method (FEM) thermal model with hundreds of elements can take significant time to simulate on a typical workstation, which makes it unsuitable for onboard reasoning during time-sensitive scenarios such as descent and landing, proximity operations, or in-space assembly. Further, the lack of fast and accurate thermal modeling drives thermal designs to be more conservative and leads to spacecraft with larger mass and higher power budgets.
The emerging paradigm of physics-informed machine learning (PIML) presents a class of hybrid modeling architectures that address this challenge by combining simplified physics models (e.g., analytical, reduced-order, and coarse mesh models) with sample-based machine learning (ML) models (e.g., deep neural networks and Gaussian processes) resulting in models which maintain both interpretability and robustness. Such techniques enable designs with reduced mass and power through onboard thermal-state estimation and control and may lead to improved onboard handling of off-nominal states, including unplanned down-time (e.g. GOES-7 \cite{bedingfield1996spacecraft}).
The PIML model or hybrid model presented here consists of a neural network which predicts reduced nodalizations (distribution and size of coarse finite difference mesh) given on-orbit thermal load conditions, and subsequently a (relatively coarse) finite-difference model operates on this mesh to predict thermal states. We compare the computational performance and accuracy of the hybrid model to a purely data-driven neural net model, and a high-fidelity finite-difference model (on a fine mesh) of a prototype Earth-orbiting small spacecraft. The PIML based active nodalization approach provides significantly better generalization than the neural net model and coarse mesh model, while reducing computing cost by up to 1.7 $\times$ compared to the high-fidelity model. 
\end{abstract}



\section{Introduction}


Airless bodies such as satellites and spacecraft in orbit are subject to extreme variations in temperature, necessitating active thermal control of sensitive components that are onboard. Conventionally, the design and planning process of such systems requires an expert in the field of thermal analysis to carefully select and tune a simplified and fast thermal model for the particular application at hand to get the best trade-off between accuracy and computational cost. Even so, such simplified models usually lead to conservative designs with sub-optimal performance in terms of mass and power consumption. 
Contrarily, using a highly accurate simulation model would alleviate these inefficiencies however at a significant increase in computational cost, which might even render the optimization infeasible.

Data-driven surrogates have been a popular alternative in such applications where compute cost is a concern, but in aerospace applications where reliability and trustworthiness are of paramount importance, the black-box nature of purely data-driven models presents an obstacle preventing their wide-spread adoption. However, there has been an increased interest in the area of physics informed machine learning as they're a class of techniques which use physics to constrain / enhance the predictions of the machine learning components.

Therefore, in this paper we explore a hybrid PIML architecture which adaptively changes the mesh size of a finite difference simulator based on the incident thermal loads to achieve the best cost to accuracy trade-offs in that particular scenario. By only changing the parameters of the simulator, this architecture preserves the integrity and interpretability of the physics model. Essentially, once fully trained, this model automates the process of an expert tuning the thermal model depending on each simulation scenario. The rest of this section briefly outlines the traditional process of spacecraft thermal analysis as well as give a brief overview of PIML methods before stating our research objectives.


\begin{figure*}[h]
\centering
\includegraphics[width=1.0\textwidth]{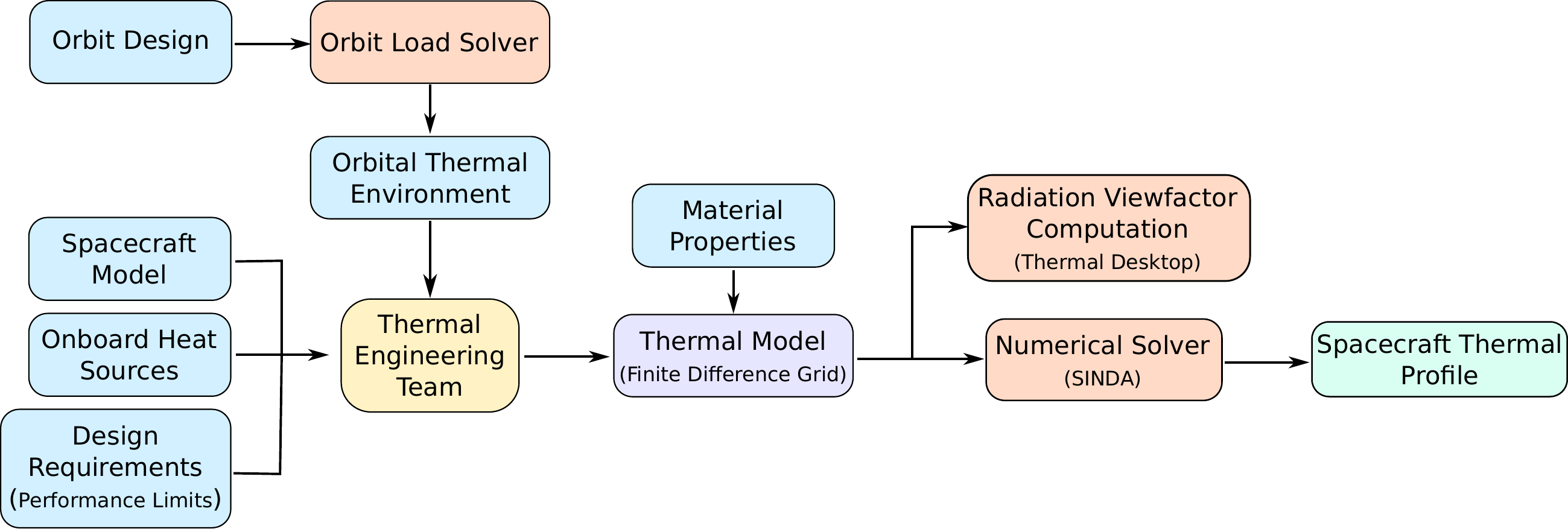}
\caption{Generic thermal modeling process for space applications }
\label{fig:thermal_modelling_process}
\end{figure*}

Figure \ref{fig:thermal_modelling_process} illustrates the thermal modelling process traditionally used for spacecraft thermal analysis. There are various inputs into the main numerical solver which are outside the scope of thermal analysis, such as orbit design and spacecraft design. However, in practice, results from the thermal analysis may feedback into the design requirements. Therefore thermal analysis needs to be performed at various levels of the space mission design process. The main bottleneck in this process requires a team of trained engineers to design a thermal model given by a grid of nodes. Engineers must consider design priorities, orbital loads, and material sensitivities when performing the nodalization step. The motivation of the present work is to assist thermal engineers in performing this crucial step with a quantified and repeatable approach.

\begin{figure*}[]
    \centering
        \centering
        \includegraphics[width=\linewidth]{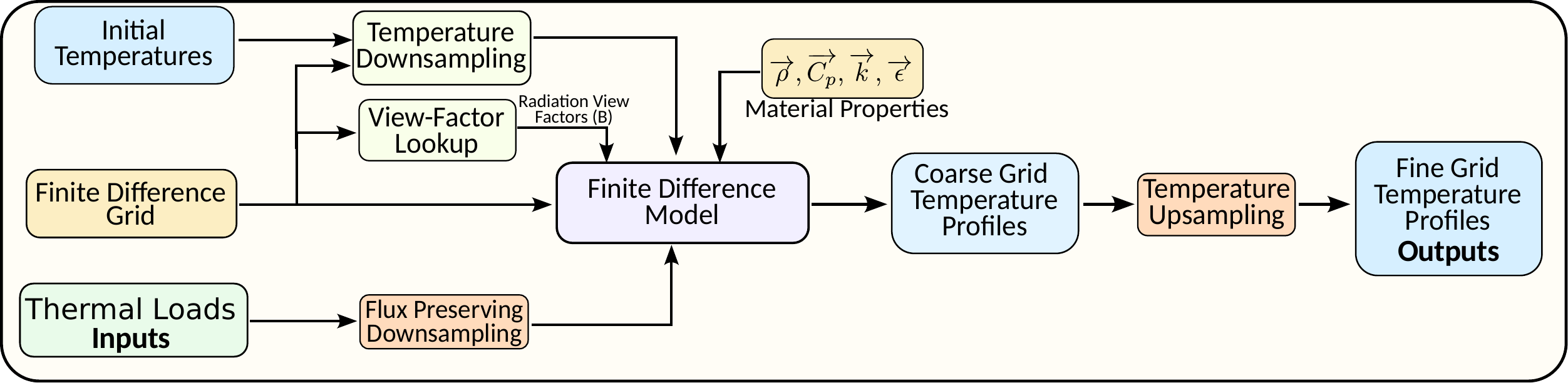}

    \caption{Architecture of the Python based auto-differentiable finite difference simulator}
    \label{fig:framework}
\end{figure*}
\subsection{Physics Informed Machine Learning (PIML) Methods}

Recent years have seen increased research interest in modelling methods that augment data-driven ML models with physics. These PIML models have additional advantages such as better generalization, extrapolation and better prediction performance on smaller datasets \cite{karniadakis2021physics}. In general, the goal of all physics informed machine learning architectures is to exploit the known physics of the problem in addition to training using data in the aim of reducing data requirements, improving predictions and/or improving model reliability. PIML architectures vary in how they combine physics and machine learning, depending on which of the above goals the architecture is tailored towards. The classic physics informed neural network (PINN) models directly incorporate physics in the loss function by using the physics partial differential equation itself as an additional loss \cite{mao2020physics, raissi2019physics, zhang2022analyses}. Since the architecture is similar to traditional neural networks these types of PIML models (or PINNs) have a similar computational cost during run time, but the added physics loss helps the PINNs have better performance in some cases \cite{cai2021physics}. Although these architectures are more ``physics-conforming'' as compared to traditional neural networks, they still are relatively black-box and provide no guarantees of physics conformity when extrapolating. 

Hybrid architectures are a class of PIML models which typically contain an embedded, fast to run physics model. They can be broadly categorized as either serial or parallel architectures. Serial architectures, shown in works such as \cite{narendra1990identification, young2017physically, nourani2009combined, singh2019pi}, typically involve sequencing data-driven models with partial physics models or utilizing them to adjust the parameters of the partial physics model. On the other hand, parallel architectures, as represented by works like \cite{javed2014robust, cheng2009fusion, karpatne2017physics}, generally consist of additive or multiplicative ensembles of partial physics and data-driven machine learning (ML) models \cite{mangili2013development}. Numerous hybrid PIML architectures have been documented in recent literature \cite{kapusuzoglu2020physics, rufa2020towards, rai2021hybrid, matei2021controlling, zhang428midphynet, chen2021hybrid, choi2021hybrid, machalek2022dynamic, lai2022intelligent, freeman2022physics, rajagopal2022physics, ankobea2022hybrid, maier2022known}, spanning diverse applications such as modeling dynamic systems, cyber-physical systems, robotic systems, flow systems, and materials behavior, among others.

The OPTMA architecture \cite{behjat2020physics}, a predecessor of the PIML architecture used in this paper, represents a physics-infused machine learning approach, combining an artificial neural network with a partial physics model to make predictions. The core mechanism involves the use of a data-driven model, specifically the transfer network, to map original inputs to those of the partial physics model, aligning its output with that of the high-fidelity model. Consequently, the training process revolves around the transfer network learning this mapping. Notably, the OPTMA model, due to its interpretable intermediate parameters, offers greater transparency compared to purely data-driven models, making it less of a black box.

However, in case of sequential hybrid-ML models, the integration of an external partial physics model introduces complexity and increases the cost of the training process. A previously employed strategy \cite{iqbal2022efficient,iqbal-tai-optma-2023} involved custom loss functions incorporating the partial physics in PyTorch \cite{Paszke_PyTorch_An_Imperative_2019} to facilitate backpropagation. While effective, this approach may not be universally applicable, especially in domains where PyTorch is sub-optimal for general-purpose scientific computing, particularly numerical methods. Therefore, we opt for Google JAX \cite{jax2018github} to automatically differentiate the partial physics model. JAX supports both forward and backward mode auto-differentiation, is compatible with Python, and seamlessly integrates with the transfer network implemented in PyTorch. This approach proves to be more versatile and scalable for developing computationally efficient PIML frameworks.
\subsection{Research Objectives:}
The overall goal of this paper is to explore the applicability of hybrid PIML techniques for spacecraft thermal simulation. To this end, we develop a PIML based \textit{adaptive} thermal simulation strategy to determine the best trade-offs between compute cost and accuracy based on the incident thermal loads. Our specific research objectives are as follows:
\begin{enumerate}
  \item Develop PIML architectures which can predict the best finite-difference nodalization depending on the incident thermal loads and which can also compensate for apriori mismatches between the data and the simulator.

  \item Develop a differentiable finite difference radiation and conduction simulator to enable the training of such hybrid models.

  \item Compare the performance of the PIML model to other pure-data driven and pure-physics baselines and test the ability of the transfer network to adapt to varying thermal loads across faces.
\end{enumerate}

The remainder of this paper is laid out as follows: In section\ref{sec:simulators} we describe the physics simulators in detail. Then in sec.\ref{sec:PIML}
we discuss the PIML architectures and detail in which cases they are applicable. In section\ref{sec:model_spacecraft} we describe the spacecraft modelling problem in detail before moving on to the results in section \ref{sec:results}. Finally, section \ref{sec:conclusion} contains our concluding remarks.

\section{Differentiable Physics Simulators}
\label{sec:simulators}

The PIML architectures proposed are tailored to compensate for the deficiencies in the simplified embedded physics model by using high-fidelity data. Therefore, in order to understand the rationale behind these architectures, we first go in detail regarding our differentiable physics simulator.
The rest of this section details the framework and structure of the differentiable embedded physics model. In this paper, we limit our simulation analysis only to the actual finite difference thermal analysis. Additional components of the overall thermal analysis notably the Monte-Carlo ray-tracing to compute the thermal loads as well as the radiation view-factors between components are not included in our differentiable physics model and by extension in the PIML models.
Instead, we use pre-computed values of the thermal loads and the radiation view-factors from the high-fidelity simulation data, the details of which are provided in the following sub-sections.

Our primary source of high-fidelity data is a Thermal Desktop model which computes the loads and radiation view-factors using ray-tracing and then solves for the temperature profile using the explicit finite difference method \cite{ozicsik2017finite}.
This high-fidelity model discretizes each dimension with 10 nodes, therefore all 2D rectangular surfaces have $10^2$ nodes while 1D surfaces have only 10 nodes. For convenience, throughout this paper, we use this shorthand whenever the physics model is mentioned. An n-node physics model contains $n^2$ nodes for 2D faces while having $n$ nodes per 1D surface.

Since the thermal desktop model is not auto-differentiable and therefore cannot be embedded into a PIML architecture which allows back-propagation, we built a similar explicit finite difference based conduction radiation solver in python using the JAX library. This makes it easily auto-differentiable and therefore amenable to be embedded into the PIML model. The following subsection details the architecture and procedure in this simplified physics model.

In our final implementation, the simplified physics model shown in Fig. \ref{fig:framework} is intended to work with varying number of nodes on each surface. However, the inputs to the simulator, such as thermal loads, radiation view-factors are defined on a 10-node mesh. Additionally, the simulator is also expected to output temperatures corresponding to the dense 10-node mesh. Therefore, an interpolation scheme is required to first down sample the inputs from the dense to the sparse grid and then up sample the temperatures from the sparse grid to the dense grid.

\subsubsection{View Factor Lookup}

The view-factor computation in the reference thermal desktop model is done using Monte Carlo ray-tracing. However, this would be extremely computationally expensive to do every time the PIML model is used. Therefore, we implement a look-up scheme where we pre-compute the view-factors on the dense 10-node mesh and then, depending on the nodalization of the coarse mesh, the view-factors of the closest dense node are assigned to each coarse node. 
This approach is reasonable in this case as the relative positions of the satellite surfaces are fixed. However, when the satellite surfaces are adjustable this approach would not work as even the dense view factors would vary.

\subsubsection{Down-sampling Thermal Loads}
During inference, the thermal loads input to the finite difference model are on a dense 10-node mesh. However, the physics model in the PIML is expected to run at a sparse nodalization. Traditional downsampling methods would not work in this case as that would change the total amount of energy input into the system which consequently would impact the outcome of the simulator. 
Therefore, we propose and implemented the following downsampling scheme such that the total flux input into any surface remains the same. The algorithm detailed below outlines the procedure for a 2D face. The procedure remains same for a 1D face, except that the area is replaced by length.

\begin{algorithm}
\caption{Flux-Preserving Load Downsampling (Rectangular Faces)}
\begin{algorithmic}[1]
\State \textbf{Input:} Dense mesh $D$, Sparse mesh $S$, Loads on Dense mesh $Q_D$
\State \textbf{Output:} Interpolated loads on sparse mesh $Q_S$
\State
\Procedure{Downsampling}{$D$, $S$, $Q_D$}
    \State \textbf{Step 1:} Compute the corners of all elements in the reference dense mesh
    \State \textbf{Step 2:} Compute the corners of all elements in the sparse mesh
    \State \textbf{Step 3:} Using the corners, compute the areas($A$)  of all elements in the reference dense mesh
    \State \textbf{Step 4:} Compute the thermal flux ($\phi_j = \dfrac{Q_D}{A}$) for each element $j$ in the dense mesh. 
    \For{Each element $i$ in the sparse mesh}
    \State \textbf{Step 5:} Compute the area of overlap($A_c$) with elements in the dense mesh 
    \State \textbf{Step 6:} Interpolated Load $Q_S[i] = \sum_{j=0}^n A_c[i] \times \phi_j $, where n is the number of dense elements a given sparse element intersects.
    \EndFor
\EndProcedure
\end{algorithmic}
\end{algorithm}
 
\subsubsection{Temperature Interpolation}

The PIML and the simplified physics models are designed to run on sparse nodalizations. But as the reference dataset is based on a denser nodalization, the predicted temperatures need to be interpolated to match the data during training and inference. In this paper, we use cubic spline interpolation to upscale to the dense nodalization. Additionally, the input temperature distribution (initial condition) is also defined on a dense nodalization and is therefore down-sampled using bi-linear or linear downsampling depending on whether the surface is 2D or 1D respectively.

\begin{figure*}[t]
    \centering
    \begin{subfigure}{\linewidth}
        \centering
        \includegraphics[width=\linewidth]{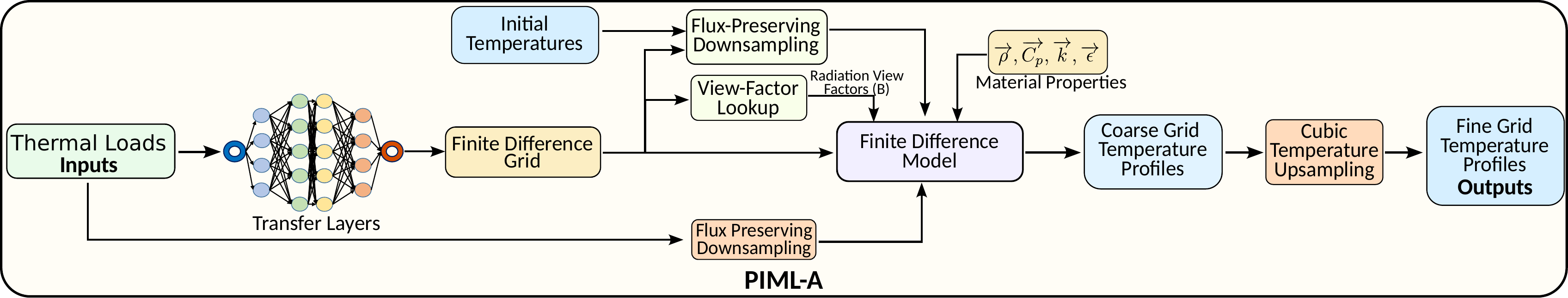}
        \caption{}
    \end{subfigure} \\
    \begin{subfigure}{\linewidth}
        \centering
        \includegraphics[width=\linewidth]{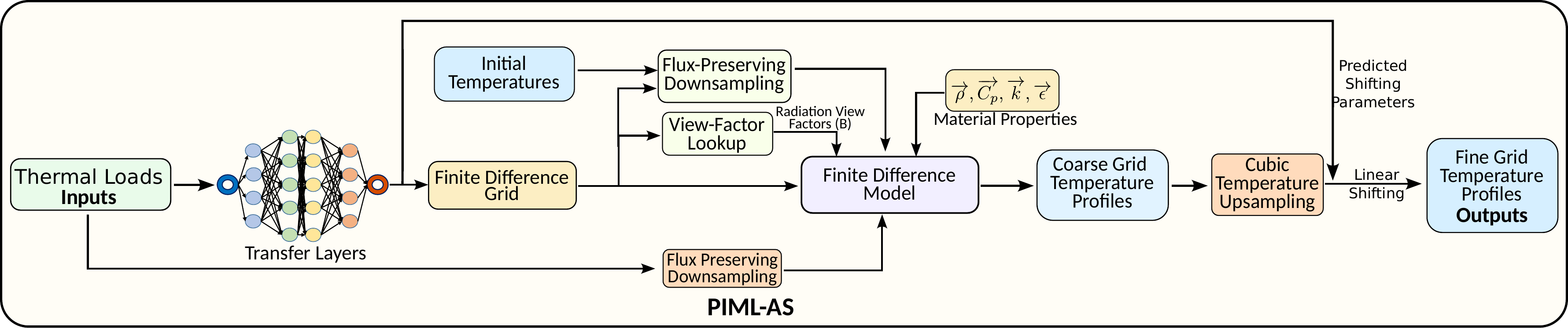}
        \caption{}
    \end{subfigure} \\

    \caption{Model Architectures of : a) PIML model designed to be trained on Python simulator data b) PIML model with output shifting, designed to be trained on Thermal Desktop simulator data }
    \label{fig:PIML framework}
\end{figure*}

\section{PIML Architectures}
\label{sec:PIML}

We developed two variants of PIML architectures shown in Fig. \ref{fig:PIML framework} that are architecturally similar but differ in what their embedded neural networks predict. Both models have a neural network which takes in the thermal loads as input and predicts the number of nodes per surface required on the spacecraft. In the PIML-AS model, the transfer layers also predict additional parameters which are used to linearly shift the output temperature predicted by the physics model. The choice between these architectures depends on the relationship between the training data and the embedded physics model.

Our initial PIML strategy embodied by PIML-A (Adapt), is designed for use when the simulator is capable of predictions in the distribution of the high-fidelity data. In our case, we use it to train the PIML model on a dataset generated by embedded simulator albeit at a denser nodalization. This model then alters the nodalization on each surface of the spacecraft based on the incident thermal loads.
Within this architecture, the input data is passed through transfer layers, which subsequently generate the number of nodes on each surface. These inputs, along with thermal loads and material properties, are fed into the finite difference simulation, which predicts the temperatures on this sparse nodalization. The final temperature values at the nodes of interest are then computed by using cubic interpolation.

Our second model, PIML-AS(Adapt and Shift), retains a similar architecture with the primary difference of having additional parameters per surface which are used for output shifting. This strategy was primarily implemented as we noticed that the temperature profiles predicted by our python simulator at the same nodalization as that of the Thermal Desktop simulator are slightly mismatched. Therefore, we task the transfer network to learn that mapping as well. This demonstrates the flexibility of these hybrid PIML architectures where an architecture can be modified to suit the available physics and data. The rest of this subsection details the losses used in training the PIML models.

\subsection{Training Losses}
\label{subsec:model_training}
Since the transfer network is essentially being trained in a semi-supervised manner, in order to constrain the network to achieve the target objective i.e adaptive nodalization, we impose two losses to force the transfer network to learn how to trade-off nodes between faces instead of simply predicting a uniform nodalization across faces.
These losses are impose first on the model predictions in the form of a mean squared error and secondly on the total number of nodes predicted so as to force the model to learn to minimize the cost as well.
The cost loss $\mathcal{L}_C$ is designed such that it keeps decreasing as a function with decreasing number of nodes, but once the total compute cost of the model exceeds the upper limit, then the penalty increases significantly and the model is forced to revert to lower nodalizations at the cost of a slightly higher mean squared error (MSE).
 The following equations list the loss functions in detail:

\begin{itemize}
  \item The mean squared error($\mathcal{L}_{m}$) to match predictions with data and
  \item A cost loss ($\mathcal{L}_{c}$) to encourage the model to use as few nodes as possible
\end{itemize}

The overall loss ($\mathcal{L}$) is then: 
\begin{equation}
\mathcal{L} = \mathcal{L}_{m} + \mathcal{L}_{c}
\label{eq:Weighted Loss}
\end{equation}

The Mean Squared Error is given as:

\begin{equation}
\label{eq:Weighted_MSE}
\begin{aligned}
\mathcal{L}_{m} &= \frac{1}{D}\sum_{d=1}^{D} w_{d} \left( \frac{1}{N} \sum_{i=1}^{N}   (T_{i,d} - \hat{T}_{i,d})^2 \right)\\
\text{where} \quad w_{d} &= 
\begin{cases} 
1 & \text{if node } d \text{ belongs to 2D surface} \\
10 & \text{if node } d \text{ belongs to 1D surface}
\end{cases}
\end{aligned}
\end{equation}

Similarly, the Cost loss is formulated as:
\begin{equation}
\begin{aligned}
    \mathcal{L}_{c} &= \frac{1}{N}\sum_{i=1}^{N} 10^{4\kappa_i}  \\
    \kappa_i &=  \dfrac{\left(\sum_{j=1}^{J} \tau_{i,j}\right) - \kappa_l}{\kappa_u-\kappa_l} -1
    \end{aligned}
\end{equation}


Where, $N$ is the total number of training samples (a training sample is from a single time point in a given orbit) and $T_{i,d}$ , $\hat{T}_{i,d}$ are the truth and predictions corresponding to node $d$ of the $i^{\rm{th}}$ training sample. $J$ is the total number of surfaces on the spacecraft. $D$ is the total number of nodes per sample, $\tau_{i,j}$ represents the nodalization (number of nodes per dimension per surface) predicted for the $j^{\rm{th}}$ face in the $i^{\rm{th}}$ sample. The term $\kappa_i$ symbolizes the normalized signed difference between the number of nodes predicted for the current sample compared to the set limit on total number of nodes per sample $\kappa_u$. 
The value of weight $w_i$ is chosen to be 10 for the 1D samples as in the output temperature vector, there are 10 times as many nodes on a rectangular surface as there are along the axis of a cylindrical surface.

Table.\ref{tab:loss_params} shows the values of the constants and limits used in the loss functions. The constraint of the max number of nodes is a proxy to set an upper bound on the maximum compute time of the model. The values of $\kappa_l$ and $\kappa_u$ chosen are equal to the total number of nodes in a 2 node and 6-node discretization respectively. Since we cap the maximum nodalization of the model with a penalty function, the losses $\mathcal{L}_m$ and $\mathcal{L}_c$  force the transfer network to learn to trade-off nodes between faces depending on the incident thermal loads instead of simply predicting a uniform higher or lower nodalization across all faces.



\begin{figure*}[h]
\begin{minipage}{0.55\linewidth}
    \centering
    \captionof{table}{Loss Parameters }
    \label{tab:loss_params}
    \begin{tabularx}{\linewidth}{>{\centering}X >{\centering}p{0.15\linewidth}}
    \toprule
     Variable    &  Value \tabularnewline
     \midrule 
     Total number of nodes ($D$) & 830 \tabularnewline
     Minimum number of nodes per simulation $\kappa_l$ & 38 \tabularnewline 
     Maximum number of nodes per simulation $\kappa_u$ & 306 \tabularnewline 
    \bottomrule
    \end{tabularx}
\end{minipage}\hfill
\begin{minipage}{0.39\linewidth}
    \centering
    \captionof{table}{Finite Difference Solver Settings}
    \label{tab:fd_settings}
    \begin{tabularx}{\linewidth}{>{\centering}X >{\centering}p{0.18\linewidth}}
    \toprule
     Simulator Property    &  Value \tabularnewline
     \midrule 
     Time stepping Scheme & Explicit \tabularnewline
     Time step value & 0.1 s \tabularnewline 
    Simulation Time & 50 s \tabularnewline 
    \bottomrule
    \end{tabularx}
\end{minipage}
\end{figure*}

\begin{figure*}[]
    \centering
    \begin{subfigure}{0.57\linewidth}
    \includegraphics[width=\linewidth]{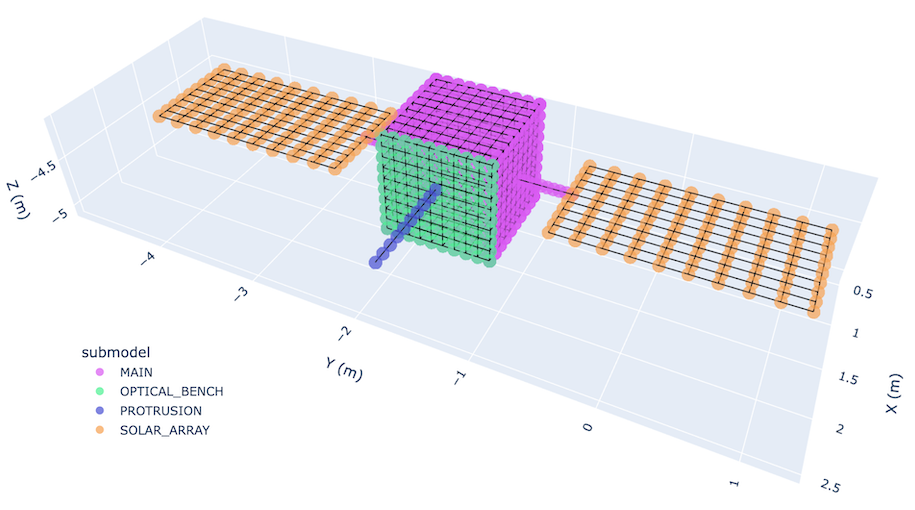}
    \caption{}
    \end{subfigure} \hfill
    \begin{subfigure}{0.4\linewidth}
    \includegraphics[width=\linewidth]{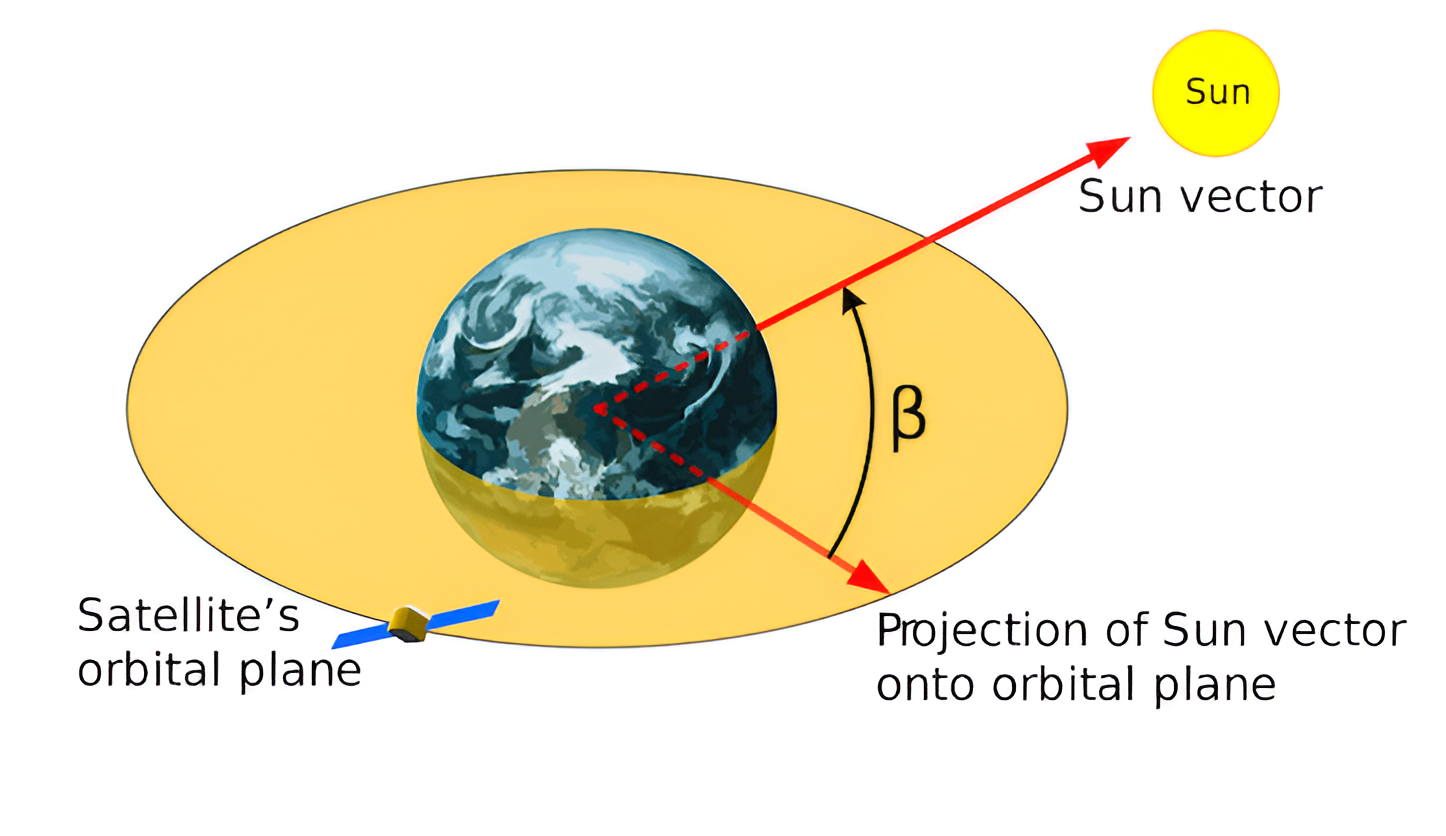}
    \caption{}
    \end{subfigure}
    \caption{Case Study Illustrations: a) Spacecraft Structure and Sub-Components b)Spacecraft in Orbit}
    \label{fig:dummy_sat}
\end{figure*}
\section{Case Study: Model Spacecraft}
\label{sec:model_spacecraft}

Figure \ref{fig:dummy_sat} shows the structure of the model spacecraft used in this analysis and also illustrates an orbit along which the thermal simulation is conducted. The model spacecraft  has two solar arrays and an Earth-pointing protrusion, which along with the optical bench side represents an instrument payload. The main spacecraft body, optical bench, solar arrays, and protrusion each have their own material properties listed in Tab. \ref{table:material_properties}. This level of detail in the spacecraft model is selected to generate a training dataset with adequate complexity. In all the simulations discussed in this paper, all the faces and components of the space craft maintain their relative positions.

In this paper, the modelling problem is poised towards predicting the temperature profile of all surfaces given the initial temperature distribution and the thermal loads. Additionally, it is also assumed that all the surfaces are insulated from each other so there is no cross-surface heat conduction possible. However, cross-surface radiation is allowed the magnitude of which is determined by the view factors.
All rectangular surfaces on the spacecraft are discretized in 2D while all cylindrical components which are discretized in 1D along their axis. The python finite difference simulator used is an explicit solver with its parameters listed in Tab. \ref{tab:fd_settings}.

\begin{table*}[!h]
\centering
\small
\caption{Spacecraft Properties}
\label{table:material_properties}
\begin{tabularx}{\textwidth}{>{\centering}p{0.08\linewidth}>{\centering}p{0.06\linewidth}>{\centering}X>{\centering}X>{\centering}X>{\centering}X>{\centering}X>{\centering}X>{\centering}X >{\centering}X}
\toprule
Component & \# of Faces & Face Type & IR Emissivity & Solar Absorptivity & Specific Heat (J/kg/K) & Thermal Conductivity (W/mK) &Density (kg/m$^3$) & Thickness (m) & Radius (m)  \tabularnewline
\midrule
Main  & 5& Rectangular&  0.82 & 0.45 &  896 & 167 &2700 & 0.01 & - \tabularnewline
Optical Bench & 1 &Rectangular& 0.92 & 0.17 &1100 & 32 & 1660 & 0.002 & - \tabularnewline
Protrusion & 1& Cylindrical & 0.92 & 0.17 & 1100&32 & 1660 & 0.002 & 0.10 \tabularnewline
Solar Array Gimbals & 2 & Cylindrical & 0.92 & 0.17 & 502&7.2 & 4429 & 0.002 & 0.05 \tabularnewline
Solar Arrays &2 &Rectangular &0.92 & 0.17 & 920& 1.7 &72 & 0.013 & - \tabularnewline
\bottomrule
\end{tabularx}
\end{table*}





\begin{table*}[h]
    \centering
    \small
    \captionof{table}{PIML and Baseline Model Parameters}
    \label{tab:models}
    \begin{tabularx}{\linewidth}{>{\centering}X >{\centering}X>{\centering}X>{\centering}p{0.15\linewidth}}
    \toprule
     Dataset & Model & Parameter   &  Value \tabularnewline
     \midrule 
     \multirow{6}{*}{Python Simulator Dataset} & \multirow{3}{*}{PIML-A} & \# of transfer parameters &  11\tabularnewline
                                               &         & \# of hidden layers & 5 \tabularnewline
                                               &         & \# of neurons per layer & 1200 \tabularnewline \cline{2-4}
                                                & \multirow{2}{*}{ANN} & \# of hidden layers& 1500 \tabularnewline        
                                               &         & \# of neurons per layer & 5 \tabularnewline \cline{2-4}
                                               & LF Model & \# of nodes per surface & 3x3 nodes for 2D discretized faces and 3 nodes per 1D discretized faces \tabularnewline \cline{2-4}
                                                & HF Model & Density of nodes in grid & 10x10 nodes for 2D discretized faces and 10 nodes per 1D discretized faces \tabularnewline \hline 
                                               
     \multirow{6}{*}{Thermal Desktop Dataset} & \multirow{3}{*}{PIML-AS} & \# of transfer parameters & 22 \tabularnewline
                                               &         & \# of hidden layers & 5 \tabularnewline
                                               &         & \# of neurons per layer & 1200 \tabularnewline \cline{2-4}
                                                & \multirow{2}{*}{ANN} & \# of hidden layers & 1500\tabularnewline        
                                               &         & \# of neurons per layers & 5 \tabularnewline \cline{2-4}
                                               & LF Model & \# of neurons per surface & 3x3 nodes for 2D discretized faces and 3 nodes per 1D discretized faces\tabularnewline \cline{2-4}
                                            & \multirow{3}{*}{HF Model} & Density of nodes in grid & 10x10 nodes for 2D discretized faces and 10 nodes per 1D discretized faces \tabularnewline
    \bottomrule
    \end{tabularx}
    \end{table*}
    
\section{Results}
\label{sec:results}

To test performance, the PIML-A is trained on data generated by our python simulator and the PIML-AS model is trained on data obtained from the thermal desktop simulator. Both PIML models are trained using the loss specified in \ref{eq:Weighted Loss}. As mentioned earlier, both datasets contain the same input loading conditions and simulation times and only differ in their temperature predictions. Additionally, on both datasets, two baseline models are also used for comparison - a purely data-driven ANN as well as a pure physics model with 3-node nodalization. The ANN is trained using the weighted MSE loss specified in \ref{eq:Weighted_MSE}.
The dataset consists of temperature and loading data on every orbit with beta angles ranging from $0^\circ$ to $90^\circ$. This complete dataset is evenly divided into train and test samples with every even numbered orbit being used for training while every odd numbered orbit is used for validation.

Table \ref{tab:models} shows the parameters of the PIML models, the 2 baseline models, and the high-fidelity physics model. The ANN and the transfer networks in the PIML models are of similar size, this was done for implementation convenience and it is likely that the PIML models will show similar performance with smaller transfer layers. However, the size of the transfer layers does not incur a significant computational cost because the primary computational bottleneck in the PIML models is the physics simulation which is orders of magnitude more expensive than the neural network.

\begin{figure*}[]
    \centering
    \begin{subfigure}{0.39\linewidth}
    \includegraphics[width=\linewidth]{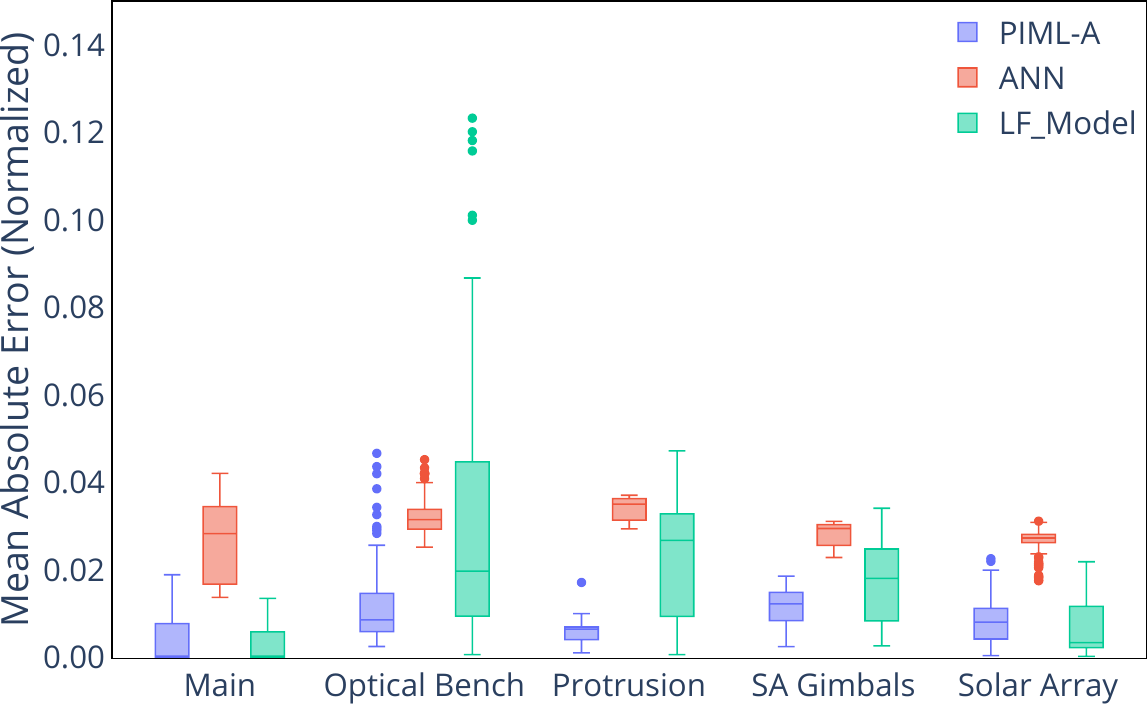}
    \caption{}
    \label{fig:rmse_py}
    \end{subfigure}\hfill
    \begin{subfigure}{0.59\linewidth}
    \includegraphics[width=\linewidth]{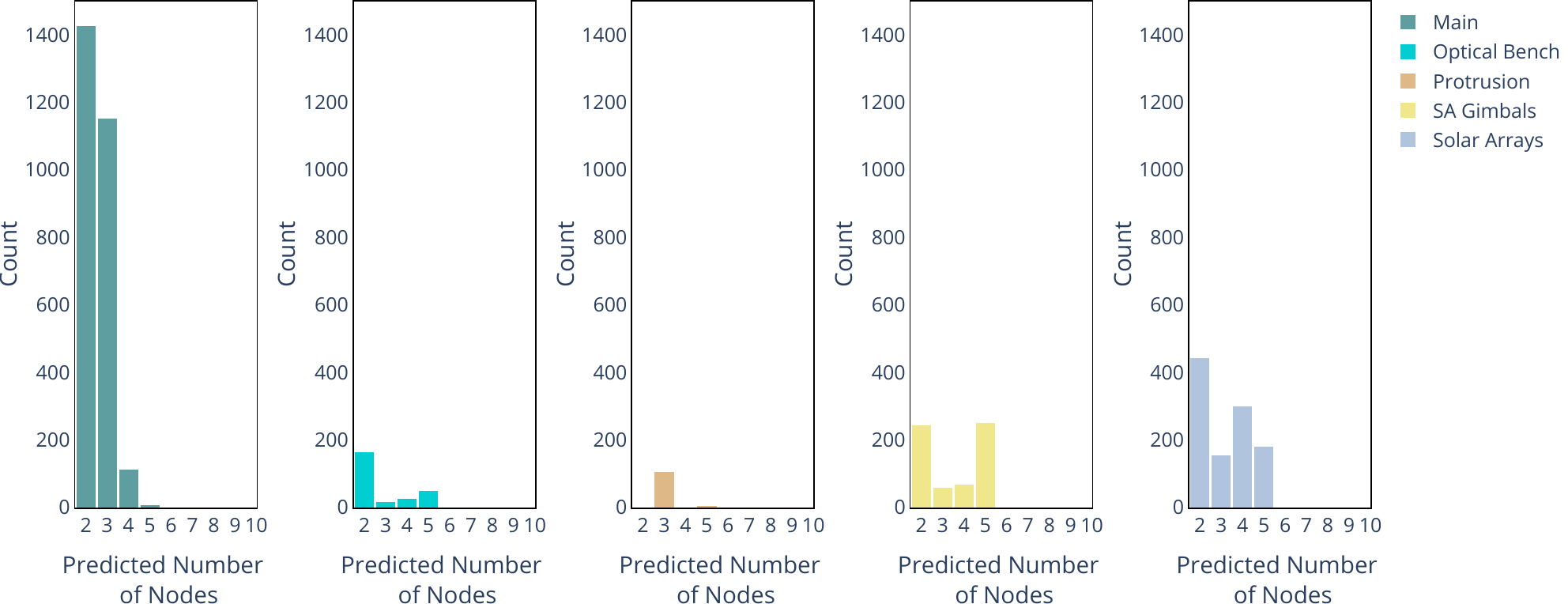}
    \caption{}
    \label{fig:node_py}
    \end{subfigure}
    \caption{Performance of PIML and baselines on Python simulator dataset: a) Mean absolute error comparison of all models b) Histogram of nodalizations predicted by the transfer network}
    \label{fig:res_py}
\end{figure*}

\begin{figure*}[h]
    \centering
    \includegraphics[width=\linewidth]{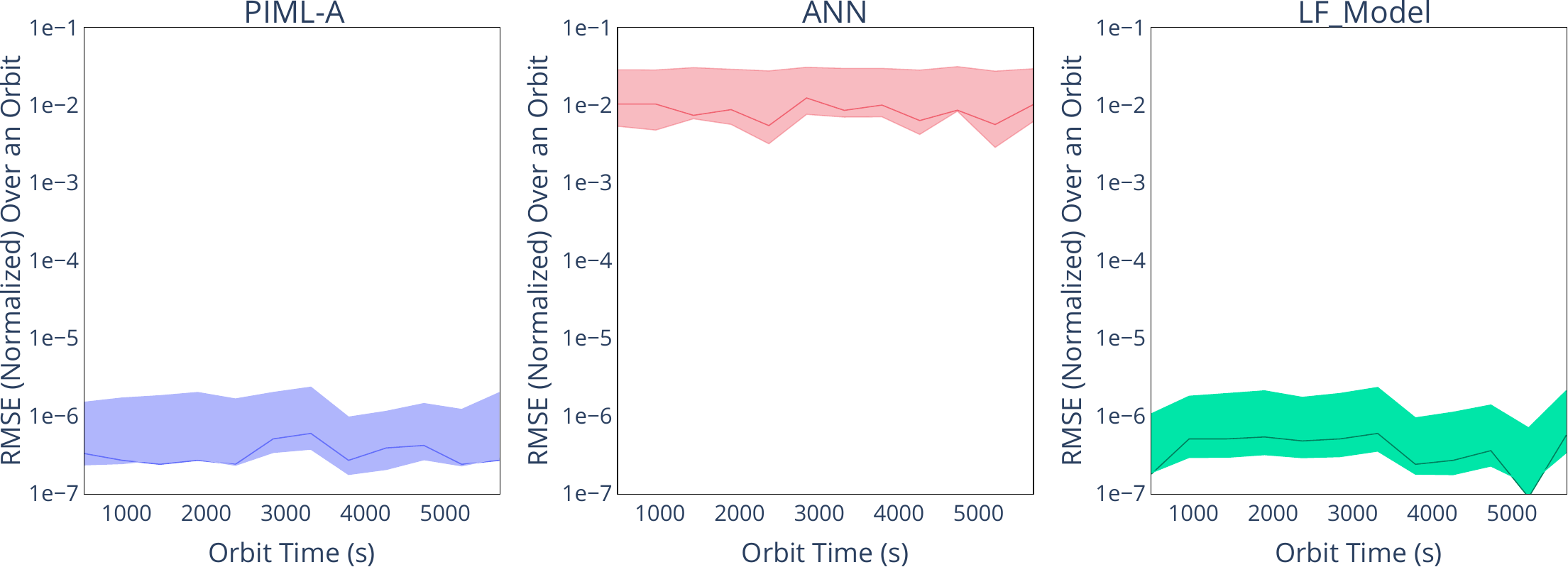}
    \caption{Illustration of error comparison across models over an orbit ($45^{\circ}$ $\beta$ angle), trained on Python simulator dataset. The median error is indicated by the solid line while the shaded region depicts the 95\% confidence interval.}
    \label{fig:orbit_py}
\end{figure*}

\subsection{Python Simulator Generated Dataset}

Figure \ref{fig:rmse_py} shows a comparison of the mean absolute error (MAE) of the PIML-A model and the 2 baseline models. From this result we can see that the PIML-A model performs better than the ANN and the baseline 3-node physics (LF) models. Further, we can see that the highest error of the low-fidelity model is on the optical bench face. This may be indicative of the fact that the loads on the optical bench face are more complex than the loads on the rest of the spacecraft, therefore the thermal gradients developed on that face cannot be represented by a simple 3-node discretization. 

Furthermore, from figure \ref{fig:node_py} we see that the transfer layers in the PIML model learn to predict a higher nodalization on the faces where the low-fidelity model produced highest errors (optical bench and protrusion faces). Simultaneously, the transfer layers also predict a lower nodalization where the low-fidelity model is already performing well. This is a very interesting result as it demonstrates the transfer layers learning to perform \textit{adaptive nodalization} i.e learning to add nodes only where required to keep the compute costs low. 

\begin{figure*}[]
    \centering
    \begin{subfigure}{0.39\linewidth}
    \includegraphics[width=\linewidth]{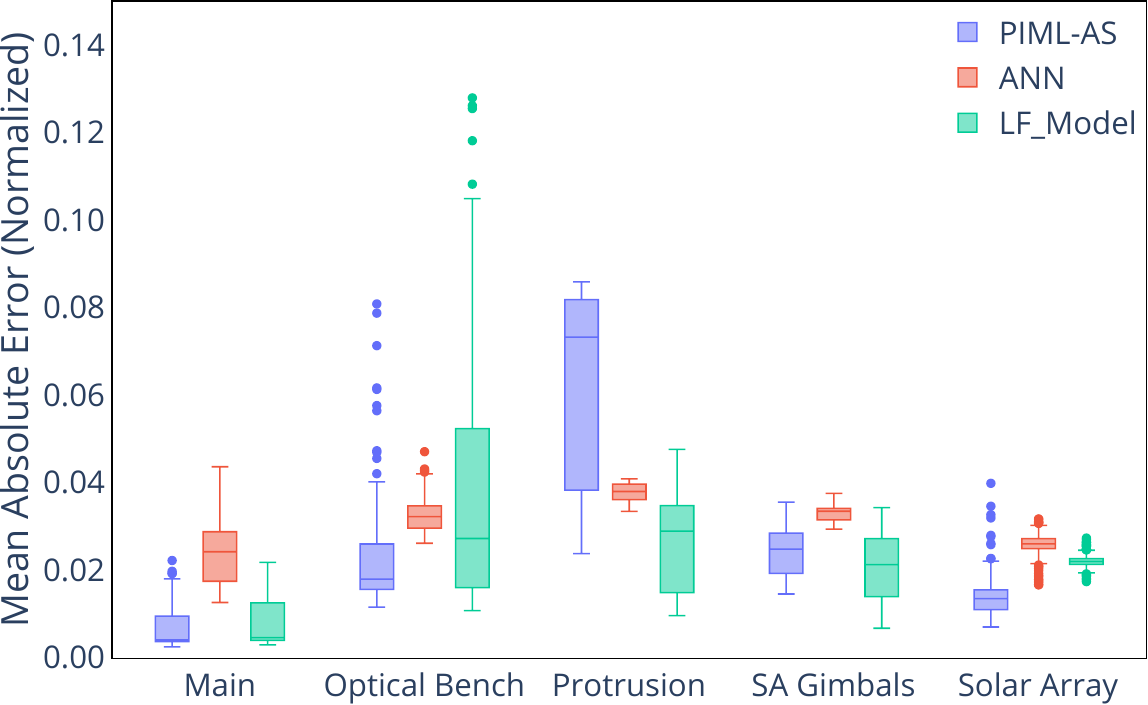}
    \caption{}
    \label{fig:rmse_td}
    \end{subfigure} \hfill
    \begin{subfigure}{0.59\linewidth}
    \includegraphics[width=\linewidth]{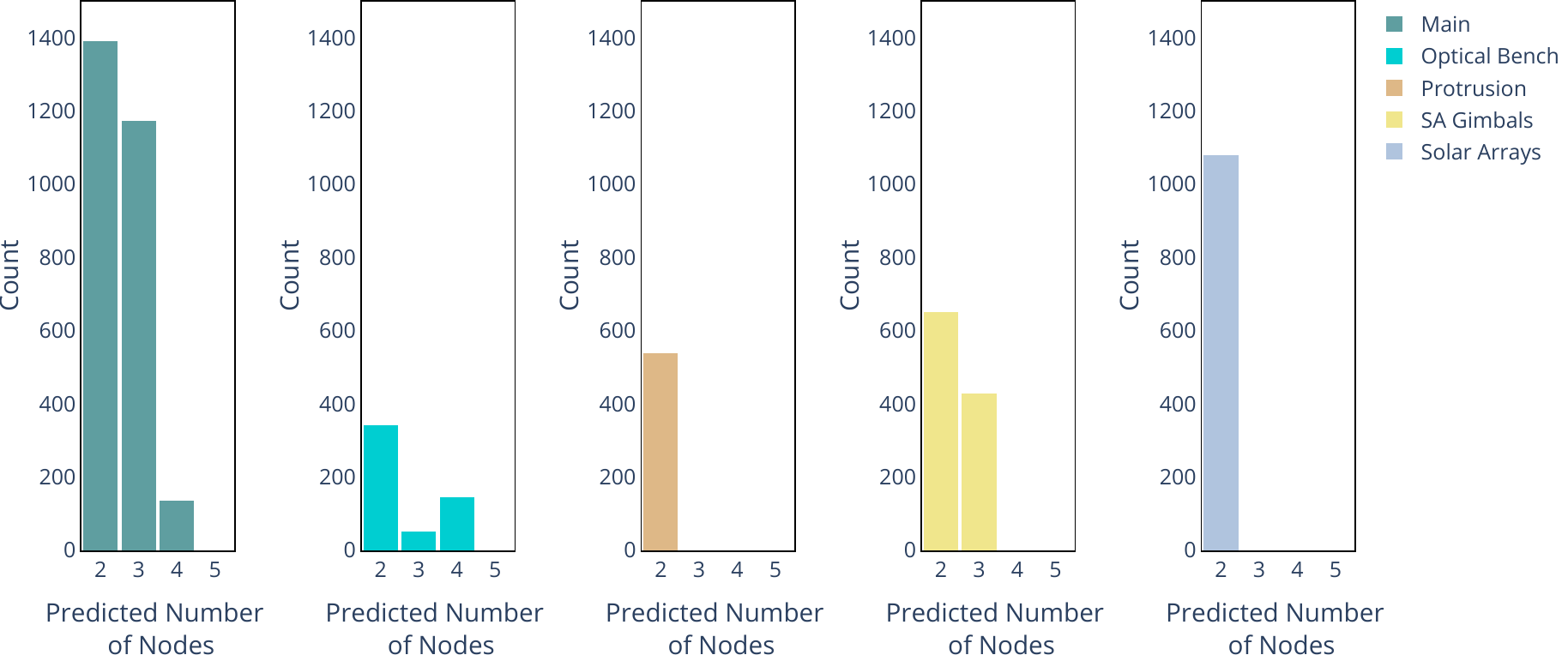}
    \caption{}
    \label{fig:node_td}
    \end{subfigure}
    \caption{Performance of PIML and baselines on Thermal Desktop dataset: a) Mean absolute error comparison of all models b) Histogram of nodalizations predicted by the transfer network}
    \label{fig:res_td}
\end{figure*}
All current predictions are at discrete time points, however, the general goal of the simulator is to be able to simulate an entire orbit. To this end, we visualize the average prediction error over the entire spacecraft for one orbit, in Figure \ref{fig:orbit_py}. Here, the results are similar to the earlier plots where the PIML and the LF model outperform the ANN model, however the variance in the error over time seems to be similar for all models.

Figures [\ref{fig:temp_profiles1},\ref{fig:temp_profiles2},\ref{fig:temp_profiles3}] in \ref{app:results_py} show the temperature contour predictions by the PIML model, the 2 baselines as well as the high-fidelity physics model for one unseen sample. The temperature contours predicted by the ANN are more noisy compared to the other models which predict much smoother contours. This behaviour is generally undesirable as it loses information about thermal gradients across a surface, which are crucial for the design of spacecraft thermal management systems. The noisy predictions can partly be attributed to the choice of ANN architecture (a Multi-Layer Perceptron (MLP)) which is sub-optimal for predicting image-like data. 

When looking at the PIML-A predictions however, we see that in faces where the low-fidelity error is low, the PIML's output matches with the low-fidelity. In cases where the low-fidelity error is high, particularly in the Optical Bench face in Figure \ref{fig:temp_profiles2}, where the PIML model's predictions are significantly different as compared to the low-fidelity model, the PIML predicts a higher nodalization and has better predictions than the low-fidelity model. This outcome, where the PIML matches the low-fidelity model where fewer nodes can capture the thermal profile and only outperforms it when higher node counts are required is a demonstration of the combination of the cost loss and the MSE loss working as intended.

\begin{table}[h]
    \centering
    \small
    \captionof{table}{Computational Time Comparision of All Models}
    \label{tab:compute_time}
    \begin{tabularx}{\linewidth}{>{\centering}X >{\centering}X >{\centering}X}
    \toprule
     Model   &  Median Run Time per Sample (s) & Median Number of Nodes Per Sample (Across All Surfaces)  \tabularnewline
     \midrule 
     PIML-A & 1.366 & 46 \tabularnewline
     PIML-AS & 1.371 & 48 \tabularnewline
     ANN & 0.003 & -\tabularnewline
     LF Model & 1.451 & 81  \tabularnewline
     HF Model(Python) & 2.256 & 830 \tabularnewline
     \tabularnewline
    \bottomrule
    \end{tabularx}
    \end{table}
    
\subsection{Thermal Desktop Generated Dataset}
\begin{figure*}
    \centering
    \includegraphics[width=\linewidth]{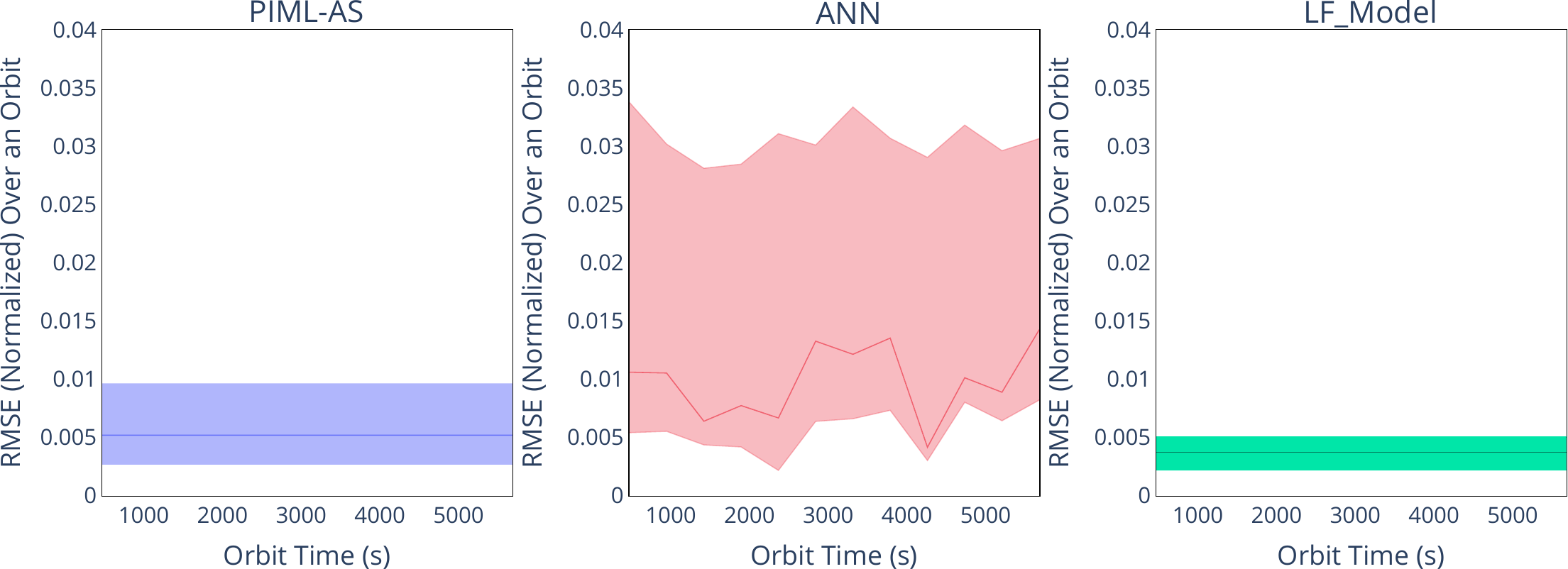}
    \caption{Illustration of error comparison across models over an orbit ($45^{\circ}$ $\beta$ angle), trained on Thermal Desktop dataset. The median error is indicated by the solid line while the shaded region depicts the 95\% confidence interval.}
    \label{fig:orbit_td}
\end{figure*}

Figure \ref{fig:rmse_td} demonstrates the performance of the PIML-AS model along with the baselines on the original 10-node thermal desktop simulator data. The results here differ significantly from the earlier dataset, with the ANN now beating the PIML on the protrusion face.
An explanation for this behaviour could be that the output mismatch between the python simulator and the thermal desktop simulator on that face cannot be captured by simple linear shifting.   To overcome this, a more complex mechanisms of compensating the simulator outputs such as a post-processing CNN at the end to correct the upscaled temperature profile may be needed. 

However, from Figure \ref{fig:node_td} we also see that the transfer network does not predict higher node counts in the protrusion. Further PIML-A predicts a higher maximum number of nodes than the PIML-AS. This could indicate that when the shifting parameters are being used, the loss gradients with respect to these parameters might be stronger than those with respect to the nodalization variables. This is because nodalization involves discrete variables, resulting in extremely shallow gradients between integer values.
For the first problem, different architectures might be a suitable solution, but for the latter issue with the gradients, the abstraction of nodalization has to be reconsidered or better approximation functions need to be formulated when converting the real number predicted by the transfer network into an integer to be used for meshing. 

The error plots over the orbit, shown in Figure \ref{fig:orbit_td} again show the higher error of the PIML and the LF models on the thermal desktop dataset as compared to the python simulator dataset. In these plots however, it is very interesting to see that the error over time for both the LF model and the PIML models remain the same. This could be because the primary source of error over the orbit is on the cylindrical faces which are being predicted at a much lower nodalization. Further experiments over several orbits need to be conducted to precisely explain this behaviour.

Figures [\ref{fig:temp_profiles1_td},\ref{fig:temp_profiles2_td},\ref{fig:temp_profiles3_td}] in \ref{app:results_td} show the temperature contours predicted by all the models trained on the Thermal Desktop Dataset. In these results the ANN has a similar performance as on the previous dataset. However, the low-fidelity model and consequently the PIML-AS models show a higher error in the predictions. Notably, the low-fidelity model predicts higher temperatures on the Solar Array surfaces as shown in Fig \ref{fig:temp_profiles2_td} while, on these same surfaces, the PIML-AS predicts lower temperatures in the same range as the high-fidelity data. This shows the effect of the shifting parameters on the predictions of the PIML-AS model. Other profiles on surfaces like Protrusion and SA Gimbals also show similar results where the temperature predictions of the PIML-AS model are in the same range as the high-fidelity data. These results also indicate to the possibility that the temperature shift between the LF model and the HF model is so high that just due to the nature of the loss function the gradients w.r.t the shifting parameters are higher and therefore the nodalization does not affect the loss as much during the initial epochs of training. Increasing the size of the transfer network and training for a greater number of epochs could result in the transfer network learning to better predict the nodalizations.

Table\ref{tab:compute_time} lists the run time per sample of all the models. The PIML models have a run-time lower than both the low and high-fidelity physics models whereas the ANN has a much smaller run time as compared to any of the physics models. This is expected as the PIML was aimed to reach the accuracy of the HF data while saving compute costs and was allowed to have surface nodalizations lower than that of the low-fidelity model. And as the PIML-AS model has a higher nodalization as compared to the PIML-A it has a marginally greater run time. 

    

\section{Concluding Remarks}
\label{sec:conclusion}
In this paper, we proposed two ways for enhancing finite difference (FD) based thermal conduction models using machine learning in terms of trade-offs between accuracy (across orbits) and computing efficiency, with applications to analyses and operations of orbiting spacecraft. While both approaches (PIML-A and PIML-AS) adaptively change the nodalization (mesh size) per spacecraft surface, PIML-AS also adapts the scaling or shifting of the output of the coarse mesh FD model to match the fine mesh high-fidelity outputs as closely as possible; both adaptations are done in response to the input thermal loads on each surface of the spacecraft. 
To enable the efficient training of these two models using backpropagation, an auto-differentiable explicit finite-difference simulator was developed in python using the JAX library. A novel flux-preserving load downsampling scheme was also proposed and implemented in order to seamlessly adapt the boundary conditions to the variable nodalization. PIML-A was trained  on a dataset generated by an open-source python FD simulator with a denser, 10-node mesh per dimension of each spacecraft surface. Similarly, PIML-AS was trained on a dataset generated by the the thermal desktop simulator, again with a 10-node sized mesh. Results are compared to two types of  baselines, a fixed nodalization coarse mesh model and purely data-driven ANNs.

When compared, the results show that the PIML models perform far superior, roughly 50\% better, than the ANNs in terms of RMSE. The PIML models also demonstrated adaptive meshing capabilities by predicting larger number of nodes on the spacecraft surfaces where the mixed coarse mesh physics model shows relatively worse performance. 
%
Further investigations however need to be conducted on the prediction differences between the open source and thermal desktop simulators, particularly on the protrusion face, to assess the differing trends in PIML performance observed with respect to either. Moreover, future work should also explore various approaches in which the real valued transfer parameters are being converted to the integer valued nodalizations for robust computation of the backpropagating gradients during training of the PIML model. Lastly, extending this work to generalize across variations in spacecraft sizing or across orbits around small bodies where sharper thermal gradients are expected will offer further evidence regarding the unique benefits of such PIML frameworks over purely physics based or purely data-drive computational approaches.

\section*{Acknowledgments}
This research was carried out at the Jet Propulsion Laboratory, California Institute of Technology and the University at Buffalo, under a contract with the National Aeronautics and Space Administration and funded through the internal Research and Technology Development program. Partial support from National Science Foundation (NSF) award CMMI 2128578 is also acknowledged. Any opinions, findings, conclusions, or recommendations expressed in this paper are those of the authors and do not necessarily reflect the views of the NASA.
\printbibliography

@book{bedingfield1996spacecraft,
  title={Spacecraft system failures and anomalies attributed to the natural space environment},
  author={Bedingfield, Keith L and Leach, Richard D},
  volume={1390},
  year={1996},
  publisher={National Aeronautics and Space Administration, Marshall Space Flight Center}
}

@ARTICLE{iqbal-tai-optma-2023,
  author={Iqbal, Rayhaan and Behjat, Amir and Adlakha, Revant and Callanan, Jesse and Nouh, Mostafa and Chowdhury, Souma},
  journal={IEEE Transactions on Artificial Intelligence}, 
  title={Auto-Differentiable Transfer Mapping Architecture for Physics-Infused Learning of Acoustic Field}, 
  year={2024},
  volume={5},
  number={3},
  pages={1132-1146},
  doi={10.1109/TAI.2023.3248561}
}

@book{ozicsik2017finite,
  title={Finite difference methods in heat transfer},
  author={{\"O}zi{\c{s}}ik, M Necati and Orlande, Helcio RB and Cola{\c{c}}o, Marcelo J and Cotta, Renato M},
  year={2017},
  publisher={CRC press}
}

@article{raissi2019physics,
  title={Physics-informed neural networks: A deep learning framework for solving forward and inverse problems involving nonlinear partial differential equations},
  author={Raissi, Maziar and Perdikaris, Paris and Karniadakis, George E},
  journal={Journal of Computational physics},
  volume={378},
  pages={686--707},
  year={2019},
  publisher={Elsevier}
}

@article{mao2020physics,
  title={Physics-informed neural networks for high-speed flows},
  author={Mao, Zhiping and Jagtap, Ameya D and Karniadakis, George Em},
  journal={Computer Methods in Applied Mechanics and Engineering},
  volume={360},
  pages={112789},
  year={2020},
  publisher={Elsevier}
}

@article{zhang2022analyses,
  title={Analyses of internal structures and defects in materials using physics-informed neural networks},
  author={Zhang, Enrui and Dao, Ming and Karniadakis, George Em and Suresh, Subra},
  journal={Science advances},
  volume={8},
  number={7},
  pages={eabk0644},
  year={2022},
  publisher={American Association for the Advancement of Science}
}

@article{cai2021physics,
  title={Physics-informed neural networks (PINNs) for fluid mechanics: A review},
  author={Cai, Shengze and Mao, Zhiping and Wang, Zhicheng and Yin, Minglang and Karniadakis, George Em},
  journal={Acta Mechanica Sinica},
  volume={37},
  number={12},
  pages={1727--1738},
  year={2021},
  publisher={Springer}
}

@inproceedings{singh2019pi,
  title={PI-LSTM: Physics-Infused Long Short-Term Memory Network},
  author={Singh, Shubhendu Kumar and Yang, Ruoyu and Behjat, Amir and Rai, Rahul and Chowdhury, Souma and Matei, Ion},
  booktitle={2019 18th IEEE International Conference On Machine Learning And Applications (ICMLA)},
  pages={34--41},
  year={2019},
  organization={IEEE}
}

@article{karpatne2017physics,
  title={Physics-guided Neural Networks (PGNN): An Application in Lake Temperature Modeling},
  author={Karpatne, Anuj and Watkins, William and Read, Jordan and Kumar, Vipin},
  journal={arXiv preprint arXiv:1710.11431},
  year={2017}
}

@article{narendra1990identification,
  title={Identification and control of dynamical systems using neural networks},
  author={Narendra, Kumpati S and Parthasarathy, Kannan},
  journal={IEEE Transactions on neural networks},
  volume={1},
  number={1},
  pages={4--27},
  year={1990},
  publisher={IEEE}
}

@phdthesis{javed2014robust,
  title={A robust \& reliable Data-driven prognostics approach based on extreme learning machine and fuzzy clustering.},
  author={Javed, Kamran},
  year={2014},
  school={Universit{\'e} de Franche-Comt{\'e}}
}

@article{young2017physically,
  title={A physically based and machine learning hybrid approach for accurate rainfall-runoff modeling during extreme typhoon events},
  author={Young, Chih-Chieh and Liu, Wen-Cheng and Wu, Ming-Chang},
  journal={Applied Soft Computing},
  volume={53},
  pages={205--216},
  year={2017},
  publisher={Elsevier}
}

@inproceedings{cheng2009fusion,
  title={A fusion prognostics method for remaining useful life prediction of electronic products},
  author={Cheng, Shunfeng and Pecht, Michael},
  booktitle={Automation Science and Engineering, 2009. CASE 2009. IEEE International Conference on},
  pages={102--107},
  year={2009},
  organization={IEEE}
}

@article{nourani2009combined,
  title={A combined neural-wavelet model for prediction of Ligvanchai watershed precipitation},
  author={Nourani, Vahid and Alami, Mohammad T and Aminfar, Mohammad H},
  journal={Engineering Applications of Artificial Intelligence},
  volume={22},
  number={3},
  pages={466--472},
  year={2009},
  publisher={Elsevier}
}

@article{behjat2020physics,
  title={A physics-aware learning architecture with input transfer networks for predictive modeling},
  author={Behjat, Amir and Zeng, Chen and Rai, Rahul and Matei, Ion and Doermann, David and Chowdhury, Souma},
  journal={Applied Soft Computing},
  volume={96},
  pages={106665},
  year={2020},
  publisher={Elsevier}
}

@inproceedings{iqbal2022efficient,
  title={Efficient Training of Transfer Mapping in Physics-Infused Machine Learning Models of UAV Acoustic Field},
  author={Iqbal, Rayhaan and Behjat, Amir and Adlakha, Revant and Callanan, Jesse and Nouh, Mostafa and Chowdhury, Souma},
  booktitle={AIAA SCITECH 2022 Forum},
  pages={0384},
  year={2022}
}

@article{mangili2013development,
  title={Development of advanced computational methods for Prognostics and Health Management in energy components and systems},
  author={MANGILI, FRANCESCA},
  year={2013},
  publisher={Italy}
}

@article{lai2022intelligent,
  title={An intelligent system for reflow oven temperature settings based on hybrid physics-machine learning model},
  author={Lai, Yangyang and Pan, Ke and Cen, Yuqiao and Yang, Junbo and Cai, Chongyang and Yin, Pengcheng and Park, Seungbae},
  journal={Soldering \& Surface Mount Technology},
  year={2022},
  publisher={Emerald Publishing Limited}
}

@article{kapusuzoglu2020physics,
  title={Physics-informed and hybrid machine learning in additive manufacturing: application to fused filament fabrication},
  author={Kapusuzoglu, Berkcan and Mahadevan, Sankaran},
  journal={Jom},
  volume={72},
  number={12},
  pages={4695--4705},
  year={2020},
  publisher={Springer}
}

@article{machalek2022dynamic,
  title={Dynamic energy system modeling using hybrid physics-based and machine learning encoder-decoder models},
  author={Machalek, Derek and Tuttle, Jake and Andersson, Klas and Powell, Kody M},
  journal={Energy and AI},
  pages={100172},
  year={2022},
  publisher={Elsevier}
}

@article{rai2021hybrid,
  title={A hybrid physics-assisted machine-learning-based damage detection using Lamb wave},
  author={Rai, Akshay and Mitra, Mira},
  journal={S{\=a}dhan{\=a}},
  volume={46},
  number={2},
  pages={1--11},
  year={2021},
  publisher={Springer}
}

@article{chen2021hybrid,
  title={Hybrid machine learning for scanning near-field optical spectroscopy},
  author={Chen, Xinzhong and Yao, Ziheng and Xu, Suheng and McLeod, Alexander S and Gilbert Corder, Stephanie N and Zhao, Yueqi and Tsuneto, Makoto and Bechtel, Hans A and Martin, Michael C and Carr, G Lawrence and others},
  journal={ACS Photonics},
  volume={8},
  number={10},
  pages={2987--2996},
  year={2021},
  publisher={ACS Publications}
}

@article{choi2021hybrid,
  title={Hybrid Machine learning and estimation-based flight trajectory prediction in terminal airspace},
  author={Choi, Hong-Cheol and Deng, Chuhao and Hwang, Inseok},
  journal={IEEE Access},
  volume={9},
  pages={151186--151197},
  year={2021},
  publisher={IEEE}
}

@inproceedings{Paszke_PyTorch_An_Imperative_2019,
author = {Paszke, Adam and Gross, Sam and Massa, Francisco and Lerer, Adam and Bradbury, James and Chanan, Gregory and Killeen, Trevor and Lin, Zeming and Gimelshein, Natalia and Antiga, Luca and Desmaison, Alban and Kopf, Andreas and Yang, Edward and DeVito, Zachary and Raison, Martin and Tejani, Alykhan and Chilamkurthy, Sasank and Steiner, Benoit and Fang, Lu and Bai, Junjie and Chintala, Soumith},
booktitle = {Advances in Neural Information Processing Systems 32},
editor = {Wallach, H. and Larochelle, H. and Beygelzimer, A. and d'Alché-Buc, F. and Fox, E. and Garnett, R.},
pages = {8024--8035},
publisher = {Curran Associates, Inc.},
title = {{PyTorch: An Imperative Style, High-Performance Deep Learning Library}},
url = {http://papers.neurips.cc/paper/9015-pytorch-an-imperative-style-high-performance-deep-learning-library.pdf},
year = {2019}
}

@software{jax2018github,
  author = {James Bradbury and Roy Frostig and Peter Hawkins and Matthew James Johnson and Chris Leary and Dougal Maclaurin and George Necula and Adam Paszke and Jake Vander{P}las and Skye Wanderman-{M}ilne and Qiao Zhang},
  title = {{JAX}: composable transformations of {P}ython+{N}um{P}y programs},
  url = {http://github.com/google/jax},
  version = {0.3.13},
  year = {2018},
}

@article{freeman2022physics,
  title={Physics-informed turbulence intensity infusion: A new hybrid approach for marine current turbine rotor blade fault detection},
  author={Freeman, Brittny and Tang, Yufei and Huang, Yu and VanZwieten, James},
  journal={Ocean Engineering},
  volume={254},
  pages={111299},
  year={2022},
  publisher={Elsevier}
}

@article{rajagopal2022physics,
  title={Physics-driven Deep Learning for PET/MRI},
  author={Rajagopal, Abhejit and Leynes, Andrew P and Dwork, Nicholas and Scholey, Jessica E and Hope, Thomas A and Larson, Peder EZ},
  journal={arXiv preprint arXiv:2206.06788},
  year={2022}
}

@article{ankobea2022hybrid,
  title={A Hybrid Physics-Based and Stochastic Neural Network Model Structure for Diesel Engine Combustion Events},
  author={Ankobea-Ansah, King and Hall, Carrie Michele},
  journal={Vehicles},
  volume={4},
  number={1},
  pages={259--296},
  year={2022},
  publisher={MDPI}
}

@article{rufa2020towards,
  title={Towards chemical accuracy for alchemical free energy calculations with hybrid physics-based machine learning/molecular mechanics potentials},
  author={Rufa, Dominic A and Macdonald, Hannah E Bruce and Fass, Josh and Wieder, Marcus and Grinaway, Patrick B and Roitberg, Adrian E and Isayev, Olexandr and Chodera, John D},
  journal={BioRxiv},
  year={2020},
  publisher={Cold Spring Harbor Laboratory}
}

@article{maier2022known,
  title={Known operator learning and hybrid machine learning in medical imaging—a review of the past, the present, and the future},
  author={Maier, Andreas and K{\"o}stler, Harald and Heisig, Marco and Krauss, Patrick and Yang, Seung Hee},
  journal={Progress in Biomedical Engineering},
  year={2022},
  publisher={IOP Publishing}
}

@article{karniadakis2021physics,
  title={Physics-informed machine learning},
  author={Karniadakis, George Em and Kevrekidis, Ioannis G and Lu, Lu and Perdikaris, Paris and Wang, Sifan and Yang, Liu},
  journal={Nature Reviews Physics},
  volume={3},
  number={6},
  pages={422--440},
  year={2021},
  publisher={Nature Publishing Group}
}

@article{matei2021controlling,
  title={Controlling Draft Interactions Between Quadcopter Unmanned Aerial Vehicles with Physics-aware Modeling},
  author={Matei, Ion and Zeng, Chen and Chowdhury, Souma and Rai, Rahul and de Kleer, Johan},
  journal={Journal of Intelligent \& Robotic Systems},
  volume={101},
  number={1},
  pages={1--21},
  year={2021},
  publisher={Springer}
}

@article{zhang428midphynet,
  title={MIDPhyNet: Memorized infusion of decomposed physics in neural networks to model dynamic systems},
  author={Zhang, Zhibo and Rai, Rahul and Chowdhury, Souma and Doermann, David},
  journal={Neurocomputing},
  volume={428},
  pages={116--129},
  publisher={Elsevier},
  year={2021}
}

\onecolumn
\appendices
\renewcommand{\thesection}{\appendixname~\Alph{section}}
\section{Representative Examples of Predicted Temperature Contours (Python Sim Dataset)}
\label{app:results_py}
\begin{figure*}[h]
    \centering
    \vspace{-0.5cm}
    \includegraphics[width=\linewidth]{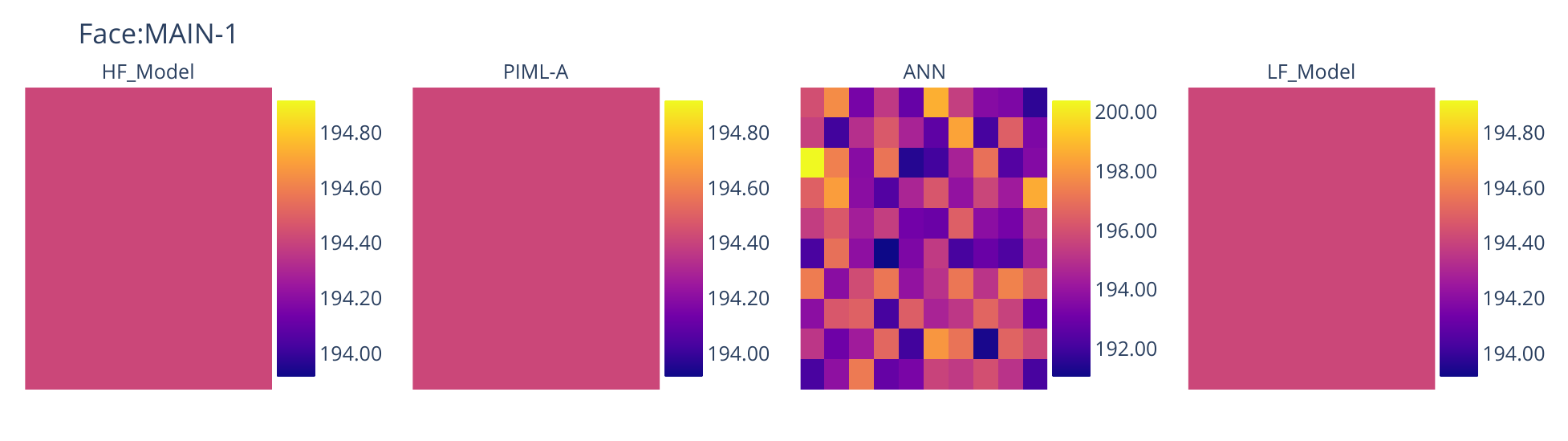}
    \vspace{-0.5cm}
    \includegraphics[width=\linewidth]{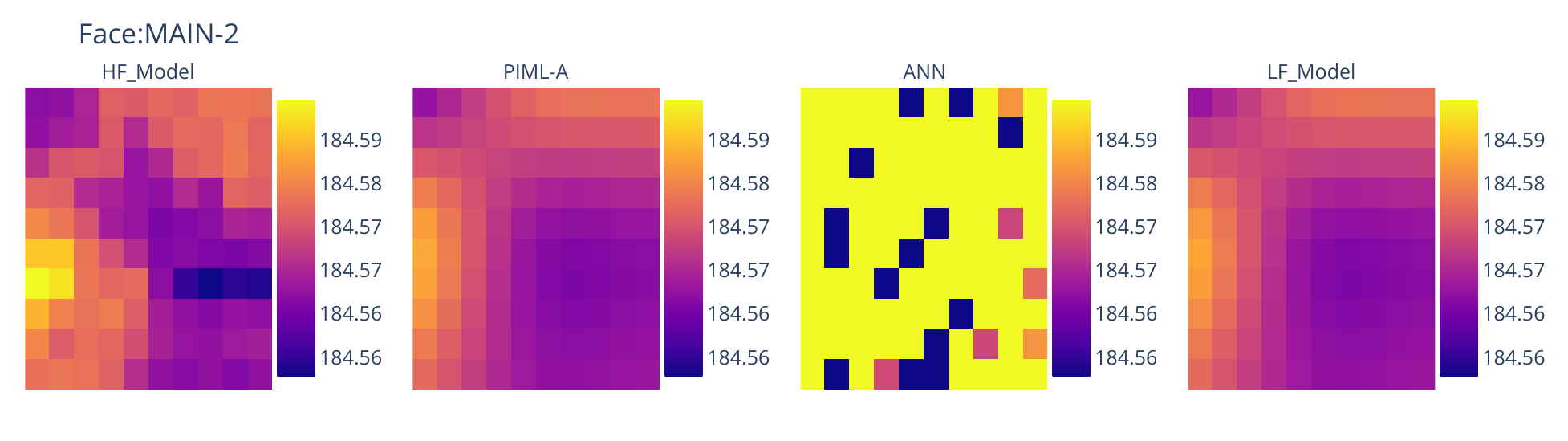}
    \vspace{-0.5cm}
    \includegraphics[width=\linewidth]{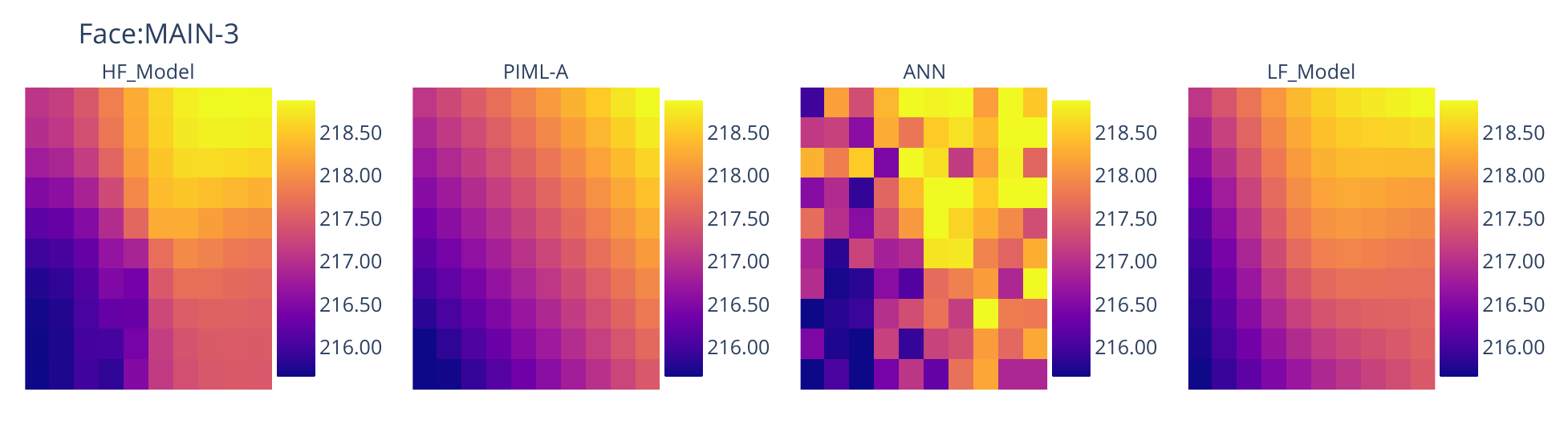}
    \vspace{-0.5cm}
    \includegraphics[width=\linewidth]{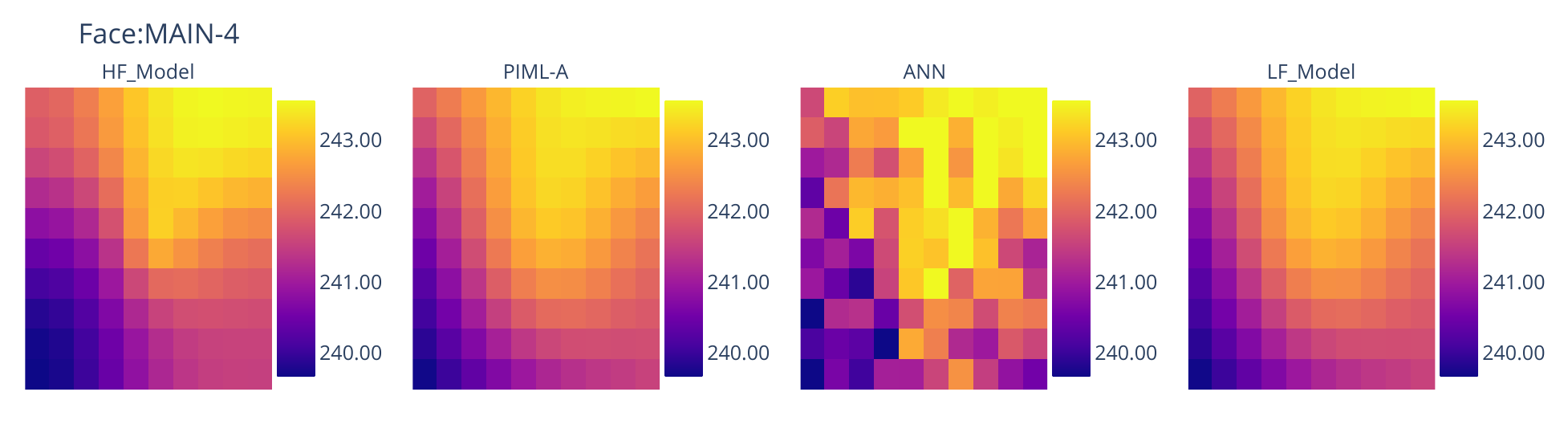}
    \caption{Temperature Predictions of All Models Across Faces Main-1 to Main-4}
    \label{fig:temp_profiles1}
    \end{figure*}
    
    \begin{figure*}[!t]
    \includegraphics[width=\linewidth]{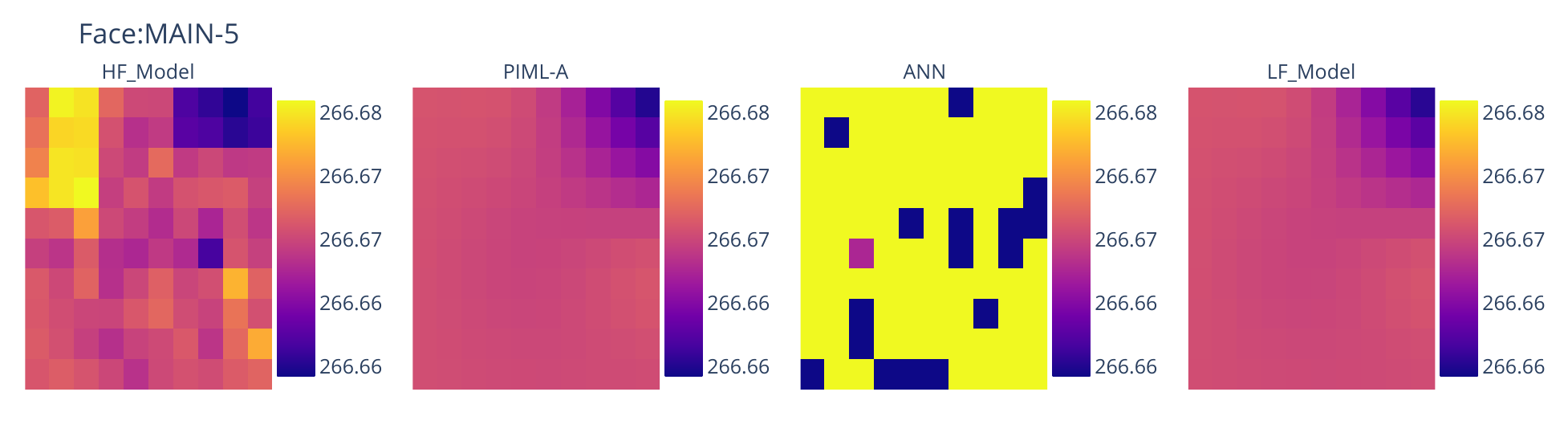}
    \includegraphics[width=\linewidth]{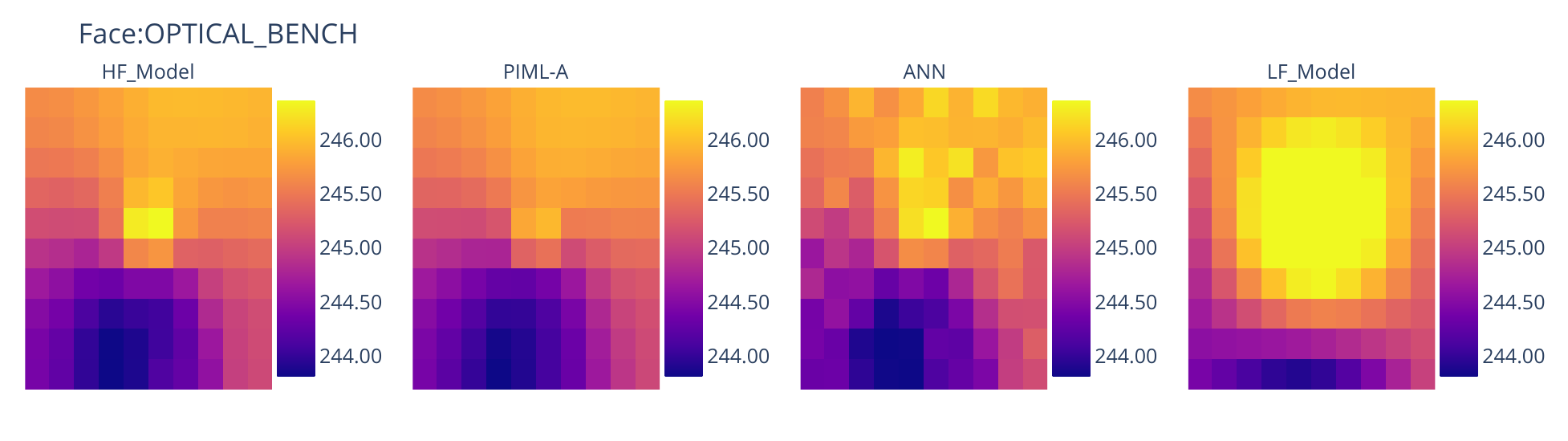}
    \includegraphics[width=\linewidth]{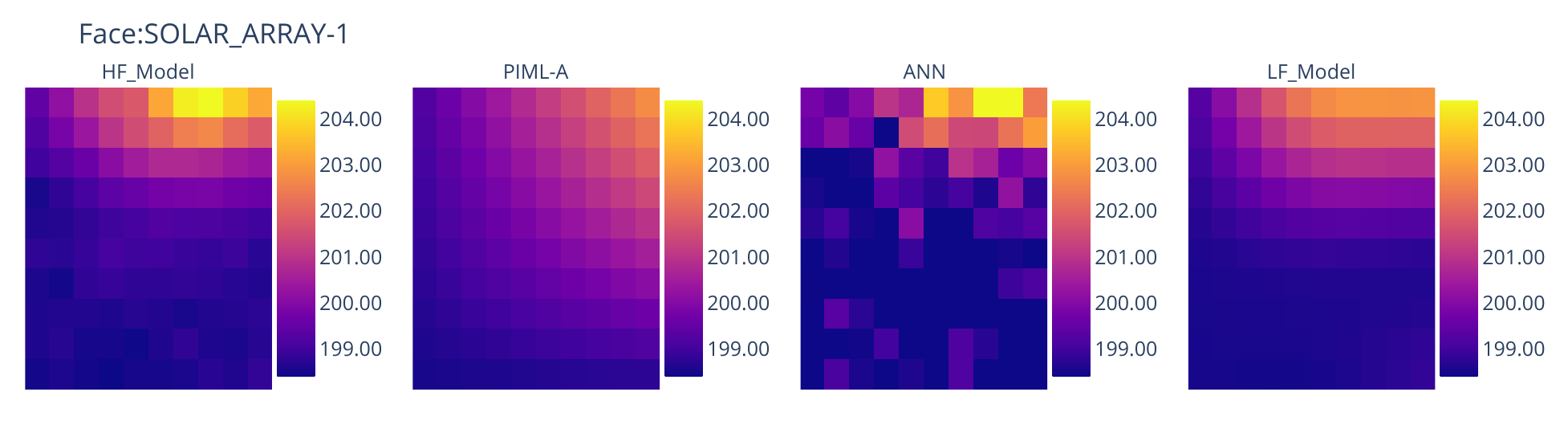}
    \includegraphics[width=\linewidth]{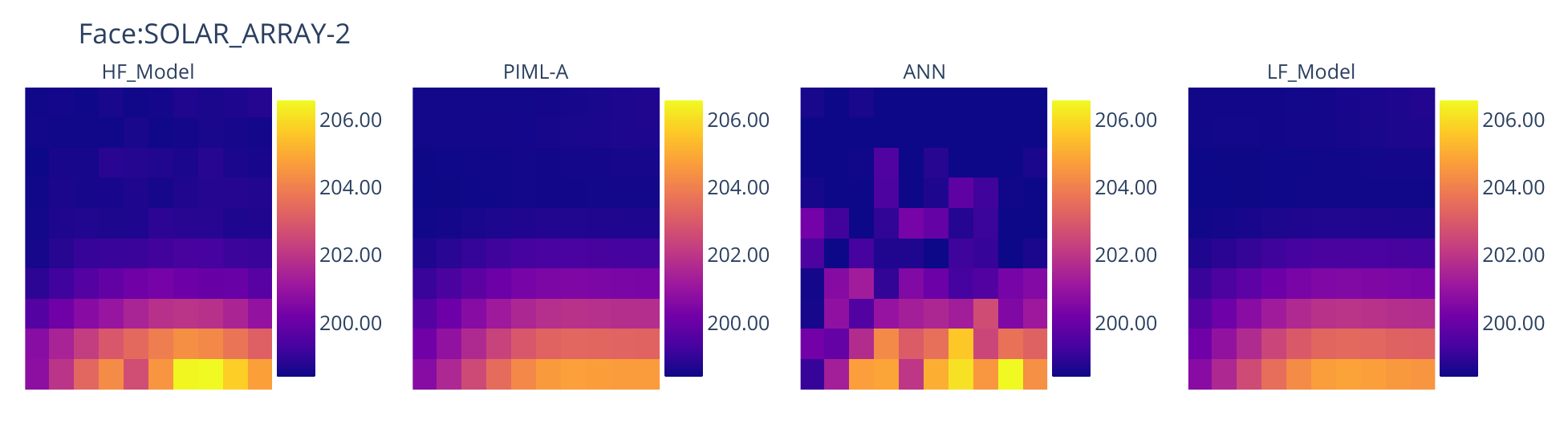}
    \caption{Temperature Predictions of All Models Across Faces: Main-5, Optical Bench, Solar Array-1, Solar Array-2}
    \label{fig:temp_profiles2}
    \end{figure*}
    \begin{figure*}[!t]
    \includegraphics[width=\linewidth]{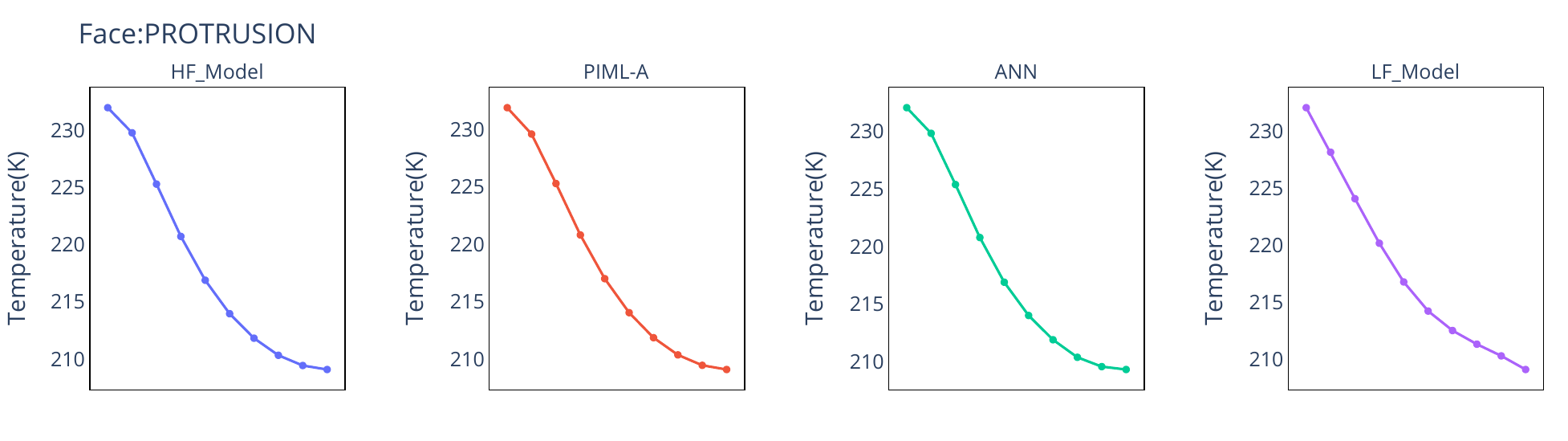}
    \includegraphics[width=\linewidth]{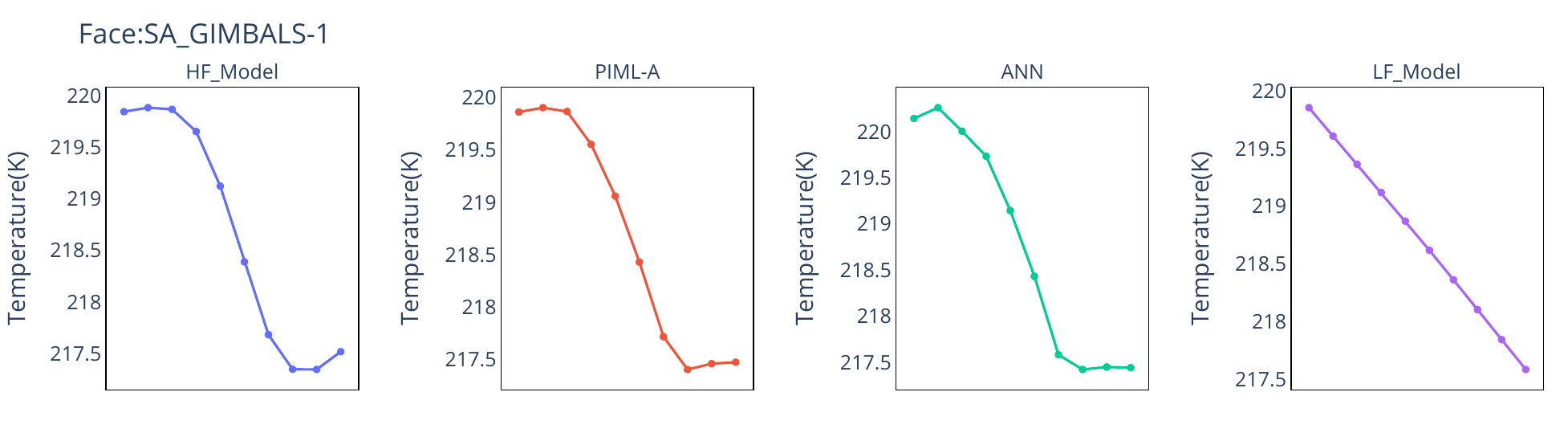}
    \includegraphics[width=\linewidth]{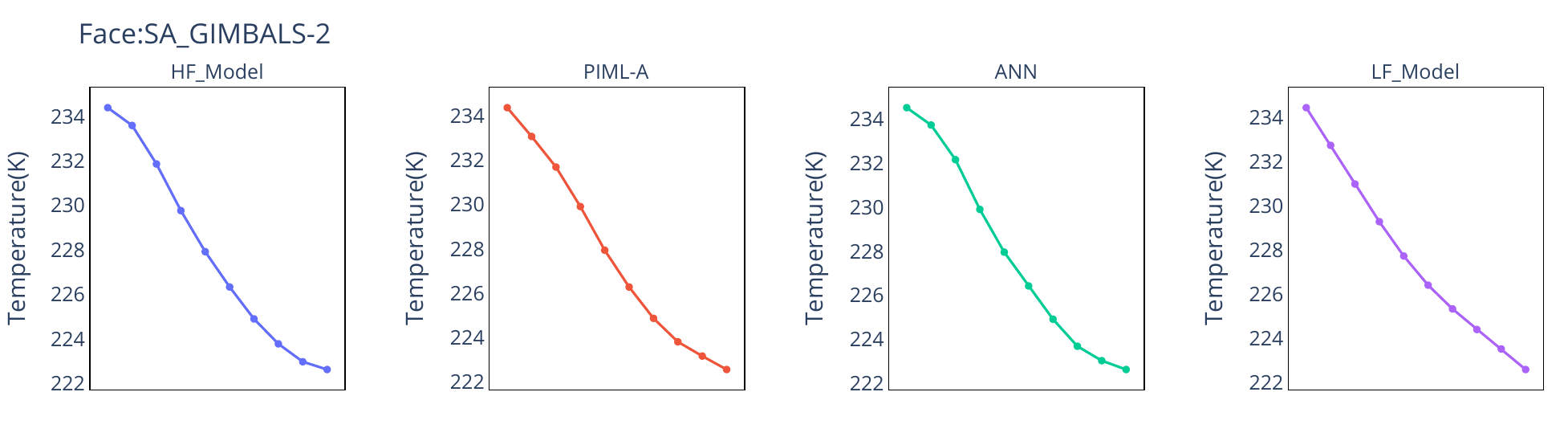}
    \caption{Temperature predictions of all models across 1D faces. Temperatures are shown on the y-axis while the x-axis represents the position of the node along the axis of the cylindrical surface.}
    \label{fig:temp_profiles3}
\end{figure*}
\clearpage

\section{Representative Examples of Predicted Temperature Contours (Thermal Desktop Dataset)}
\label{app:results_td}
\begin{figure*}[h]
    \centering
    \vspace{-0.5cm}
    \includegraphics[width=\linewidth]{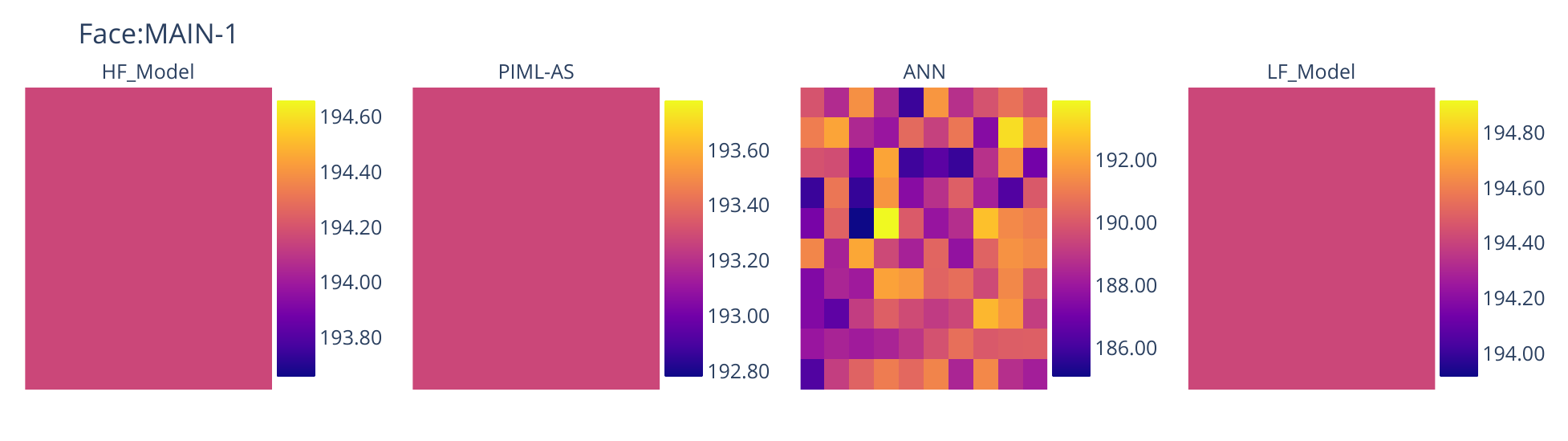}
    \vspace{-0.5cm}
    \includegraphics[width=\linewidth]{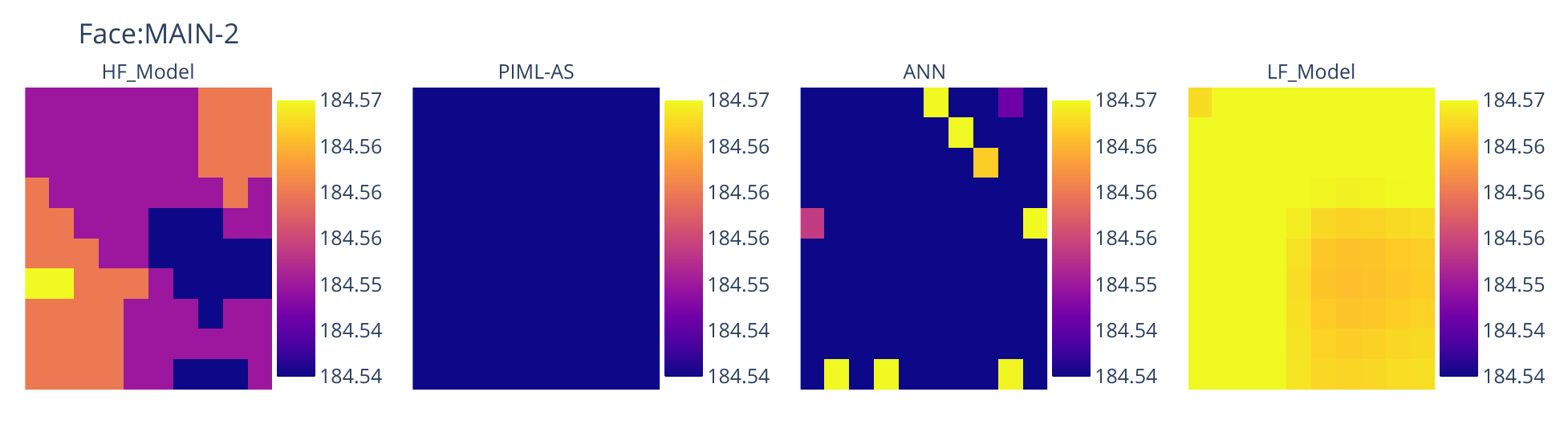}
    \vspace{-0.5cm}
    \includegraphics[width=\linewidth]{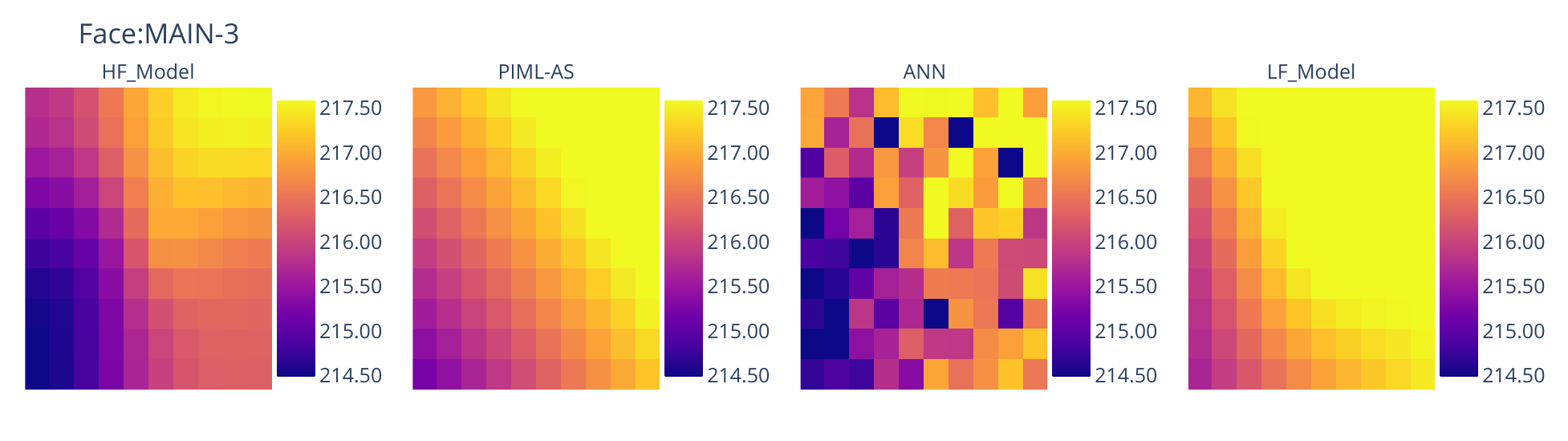}
    \vspace{-0.5cm}
    \includegraphics[width=\linewidth]{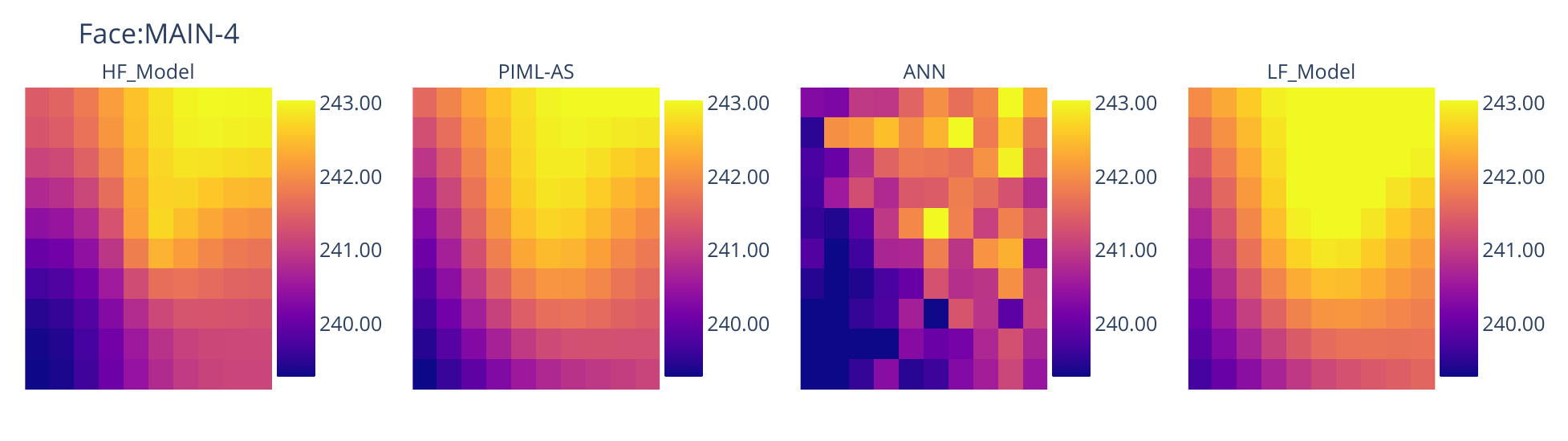}
    \caption{Temperature Predictions of All Models Across Faces Main-1 to Main-4}
    \label{fig:temp_profiles1_td}
    \end{figure*}
    
    \begin{figure*}[!t]
    \includegraphics[width=\linewidth]{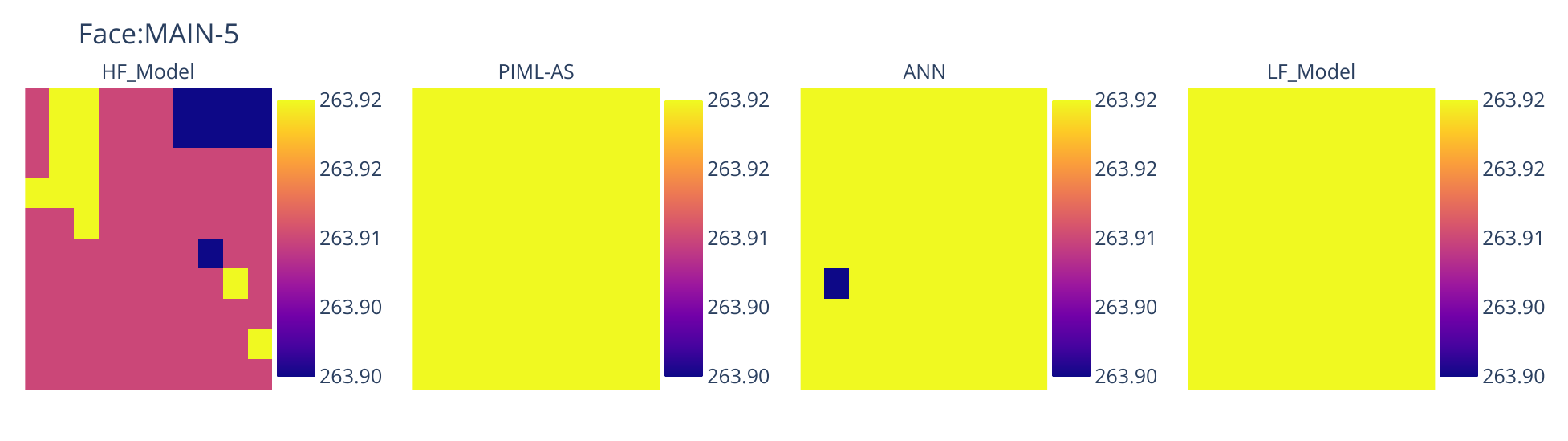}
    \includegraphics[width=\linewidth]{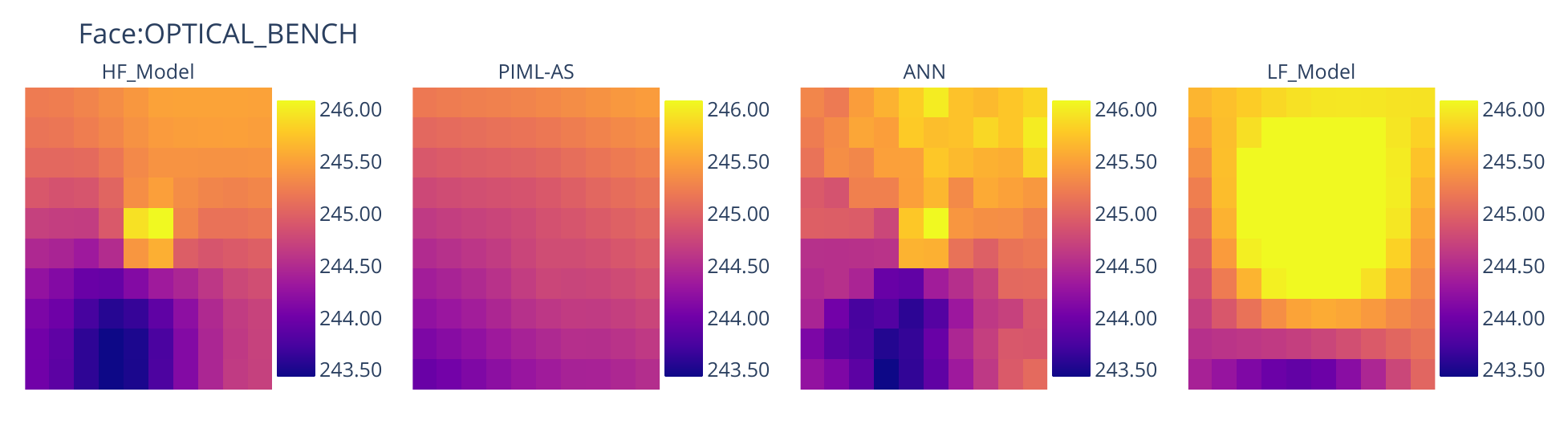}
    \includegraphics[width=\linewidth]{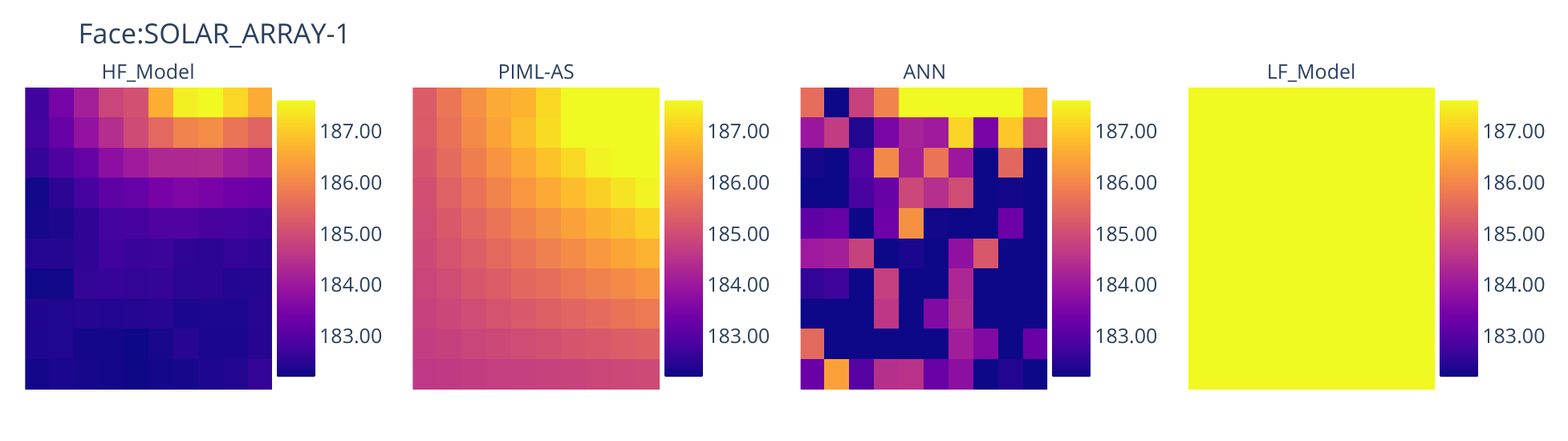}
    \includegraphics[width=\linewidth]{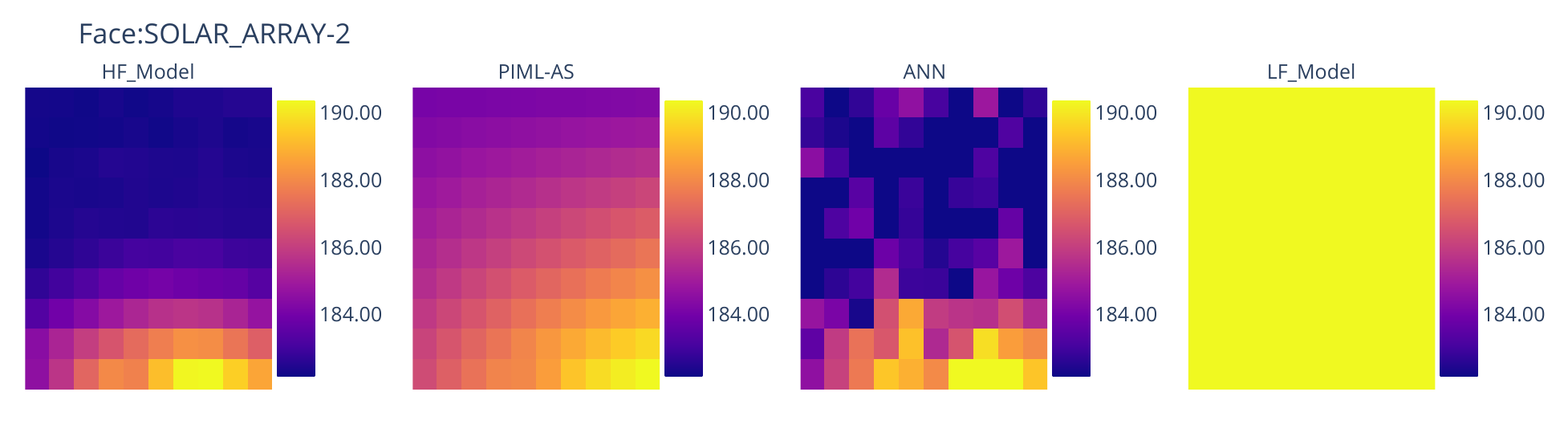}
    \caption{Temperature Predictions of All Models Across Faces: Main-5, Optical Bench, Solar Array-1, Solar Array-2}
    \label{fig:temp_profiles2_td}
    \end{figure*}
    \begin{figure*}[!t]
    \includegraphics[width=\linewidth]{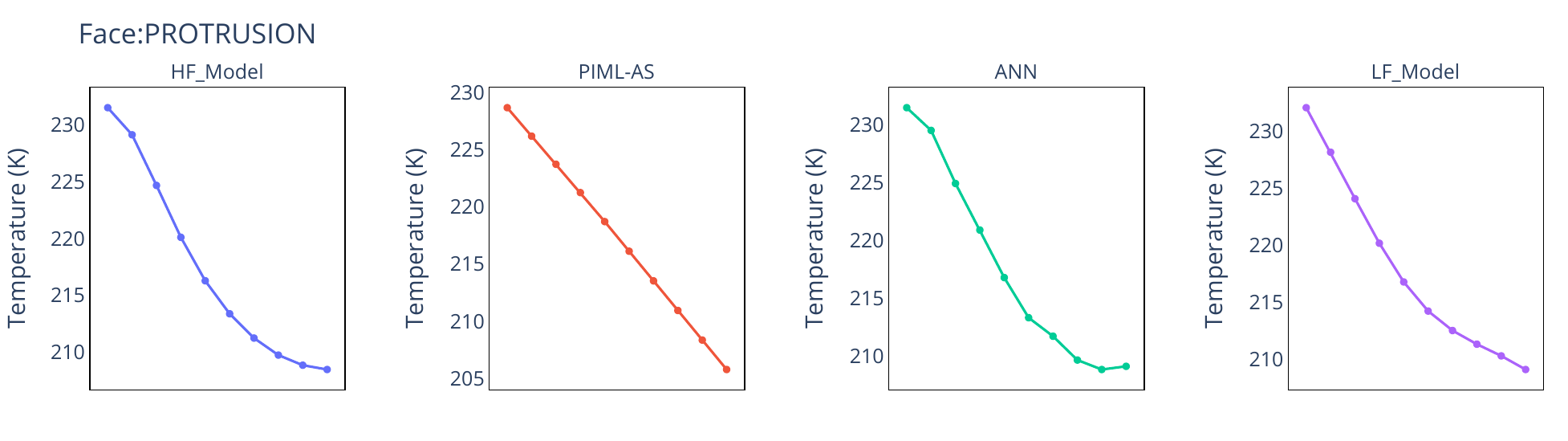}
    \includegraphics[width=\linewidth]{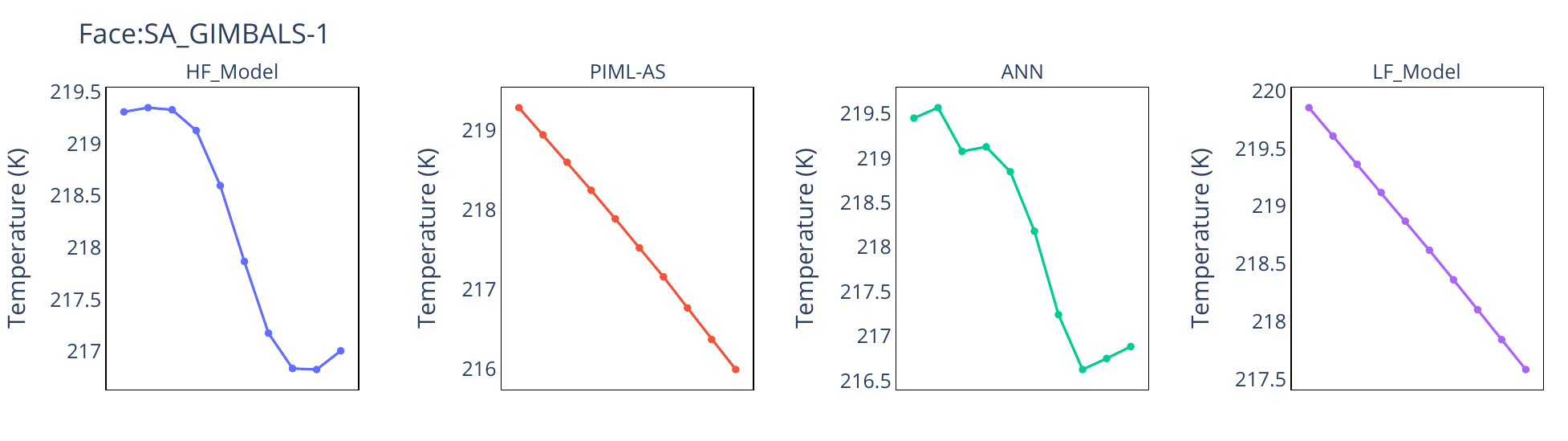}
    \includegraphics[width=\linewidth]{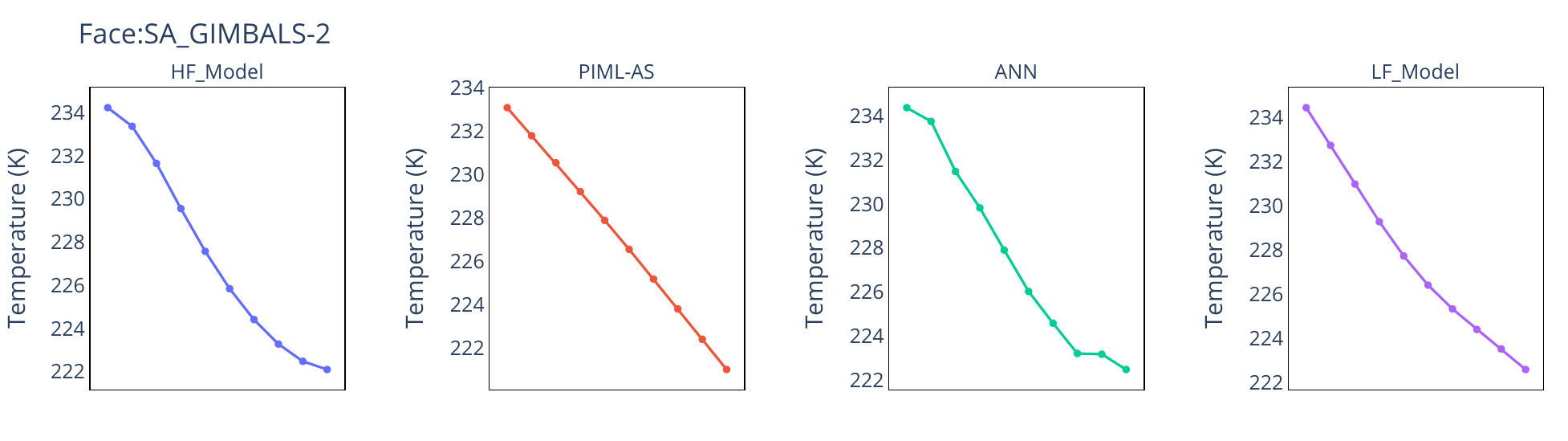}
    \caption{Temperature predictions of all models across 1D faces. Temperatures are shown on the y-axis while the x-axis represents the position of the node along the axis of the cylindrical surface.}
    \label{fig:temp_profiles3_td}
\end{figure*}
\clearpage

\end{document}